\documentclass{article} 
\usepackage{iclr2025_conference}
\usepackage{csquotes}

\usepackage{amsmath,amsfonts,bm}

\def\eqref#1{equation~\ref{#1}}

\def\1{\bm{1}}

\DeclareMathAlphabet{\mathsfit}{\encodingdefault}{\sfdefault}{m}{sl}
\SetMathAlphabet{\mathsfit}{bold}{\encodingdefault}{\sfdefault}{bx}{n}

\usepackage{soul}
\usepackage{hyperref}
\usepackage{url}
\usepackage{amssymb} 
\usepackage{graphicx}
\usepackage{algorithm}
\usepackage{algpseudocode}
\usepackage{amsmath}
\usepackage{cleveref}
\usepackage{amsthm}
\usepackage{subcaption}
\usepackage{booktabs}
\usepackage{multirow}

\newcommand{\methodname}{\textsc{WIND}}

\usepackage{subcaption}
\usepackage{geometry}
\usepackage{float}
\usepackage{thm-restate}
\usepackage{etoc}
\usepackage[toc,page,header]{appendix}
\usepackage{hyperref}
\usepackage{appendix}
\usepackage{minitoc} 
\usepackage{tocloft}
\usepackage{hyperref}
\usepackage[toc,page,header]{appendix}
\usepackage{minitoc}
\usepackage{wrapfig}  
\usepackage{booktabs} 
\usepackage{lipsum}   
\usepackage{color}

\title{
Hidden in the Noise: Two-Stage Robust \\ Watermarking for Images}

\author{Kasra Arabi, Benjamin Feuer, R. Teal Witter, Chinmay Hegde, Niv Cohen \\
New York University
}

\iclrfinalcopy 
\everypar{\looseness=-1}
\begin{document}


\maketitle
\begin{abstract}
As the quality of image generators continues to improve, deepfakes become a topic of considerable societal debate. Image watermarking allows responsible model owners to detect and label their AI-generated content, which can mitigate the harm. Yet, current state-of-the-art methods in image watermarking remain vulnerable to forgery and removal attacks. 
{This vulnerability occurs in part because watermarks distort the distribution of generated images, unintentionally revealing information about the watermarking techniques.}

In this work, we first demonstrate a distortion-free watermarking method for images, based on a diffusion model's initial noise.
However, detecting the watermark requires comparing the initial noise reconstructed for an image to all previously used initial noises. 
To mitigate these issues, we propose a two-stage watermarking framework for efficient detection. During generation, we augment the initial noise with generated Fourier patterns to embed information about the group of initial noises we used. For detection, we (i) retrieve the relevant group of noises, and (ii) search within the given group for an initial noise that might match our image. This watermarking approach achieves state-of-the-art robustness to forgery and removal against a large battery of attacks.
The project code is available at \url{https://github.com/Kasraarabi/Hidden-in-the-Noise}.
\end{abstract}


\section{Introduction}
\label{sec:introduction}

Generative AI is capable of synthesizing high-quality images indistinguishable from real ones. This capability can be used to deliberately deceive. Fake image generation have the potential to cause severe societal harms through the spread of confusion and misinformation \citep{Peebles2022DiT, esser2024scalingrectifiedflowtransformers, chen2024pixartsigma, ramesh2021zeroshottexttoimagegeneration}. In addition, owners of different models and images may want to control the spread of their derivatives for copyright reasons and to safeguard their intellectual property.
One way to mitigate these harms is model watermarking. The study of watermarking has a rich history and has recently been adopted for AI-generated content \citep{pun1997rotation, langelaar2000watermarking, craver1998resolving}. For an extended discussion of recent work in this area, we direct the reader to \Cref{sec:related-works}. Unfortunately, most current image watermarking methods are not robust to watermark removal attacks utilizing image diffusion generative models \citep{zhao2023invisibleimagewatermarksprovably}.

{Recently, new watermarking methods utilize the inversion property of DDIM to achieve more robust watermarking \mbox{\citep{wen2023tree,ci2024ringid,yang2024gaussian}}. 
These methods embed patterns in a diffusion model's initial noise and then detect them in the noise pattern reconstructed from the generated image. This technique provides strong robustness against various attacks, making it effective at resisting watermark removal attempts. 
Yet, these prior methods are themselves vulnerable to new types of attacks. Tree-Ring~\mbox{\cite{wen2023tree}} adds a pattern to the initial noise, making it distinct from a random Gaussian initial noise in a way that an attacker can detect \mbox{\citep{yang2024steganalysisdigitalwatermarkingdefense}}. This may enable forgery attacks, aimed at applying the watermark without the owner's permission. Such attacks are often even more concerning than removal attacks, as they can cause severe damage to model owners if their watermark is associated with illegal content.}

\begin{figure}[t!]
    \centering
    \includegraphics[width=0.9\linewidth]{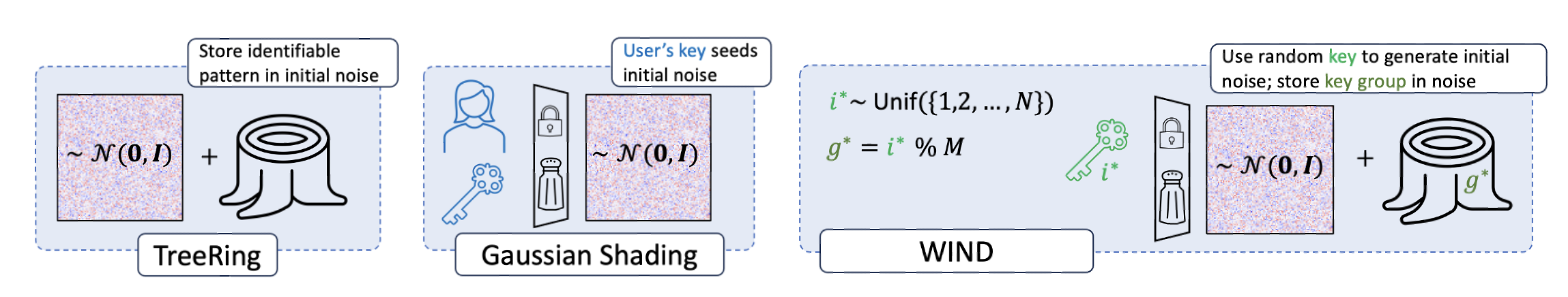}
    \caption{Related watermarking methods. Tree-Ring embeds an identifiable pattern into the initial noise \cite{wen2023tree}. Gaussian Shading uses a user-specific key to seed the initial noise \cite{yang2024gaussian}. Our method, WIND, samples a random key from $N$ options to seed the initial noise. In order to speed up detection time, the key's group is then embedded into the initial noise.}
    \label{fig:graphic_comparison}
\end{figure}

{Therefore, there is a need for image watermarking methods that generate images that are not distinct from non-watermarked images (to anyone but the model owner). As suggested by previous works, since the model already takes random noise as initialization, we may initialize it with a pseudo-random noise pattern that we can detect later~\mbox{\citep{yang2024gaussian,kuditipudi2023robust}}. Namely, reconstructing an approximation of the \textit{initial noise} used in the diffusion process from a given image allows the detection of the noise pattern used by the model. Although this reconstructed noise is not completely identical to the initial noise, it is much more similar to the initial noise than it is to other randomly distributed noise patterns. Thus, it can serve as a watermark that can be identified in the generated images. A comparison of these watermarking approaches is illustrated in \mbox{\Cref{fig:graphic_comparison}} (see also \mbox{\Cref{app:additional_disc}} for similar ideas used in previous works).}

{While using a pseudo-random initial noise does not distort the distribution of individual generated images, it may carry information about the watermark when groups of images are examined together. Specifically, works such as~\mbox{\cite{yang2024gaussian}} embeds the watermark in an initial noise such that the resulting generated image comes from the same distribution as non-watermarked images. Yet, when many images generated from the same noise pattern are examined together, the correlation between them may expose that they are not distortion-free as a set. E.g., the average of many similarly-watermarked images may differ from the average of non-watermarked ones~\mbox{\citep{yang2024steganalysisdigitalwatermarkingdefense}}. A natural solution to this distortion of sets is to use many different initial noises for each watermark we deploy.}

\begin{figure}[t]
    \centering
    \includegraphics[width=0.9\linewidth]{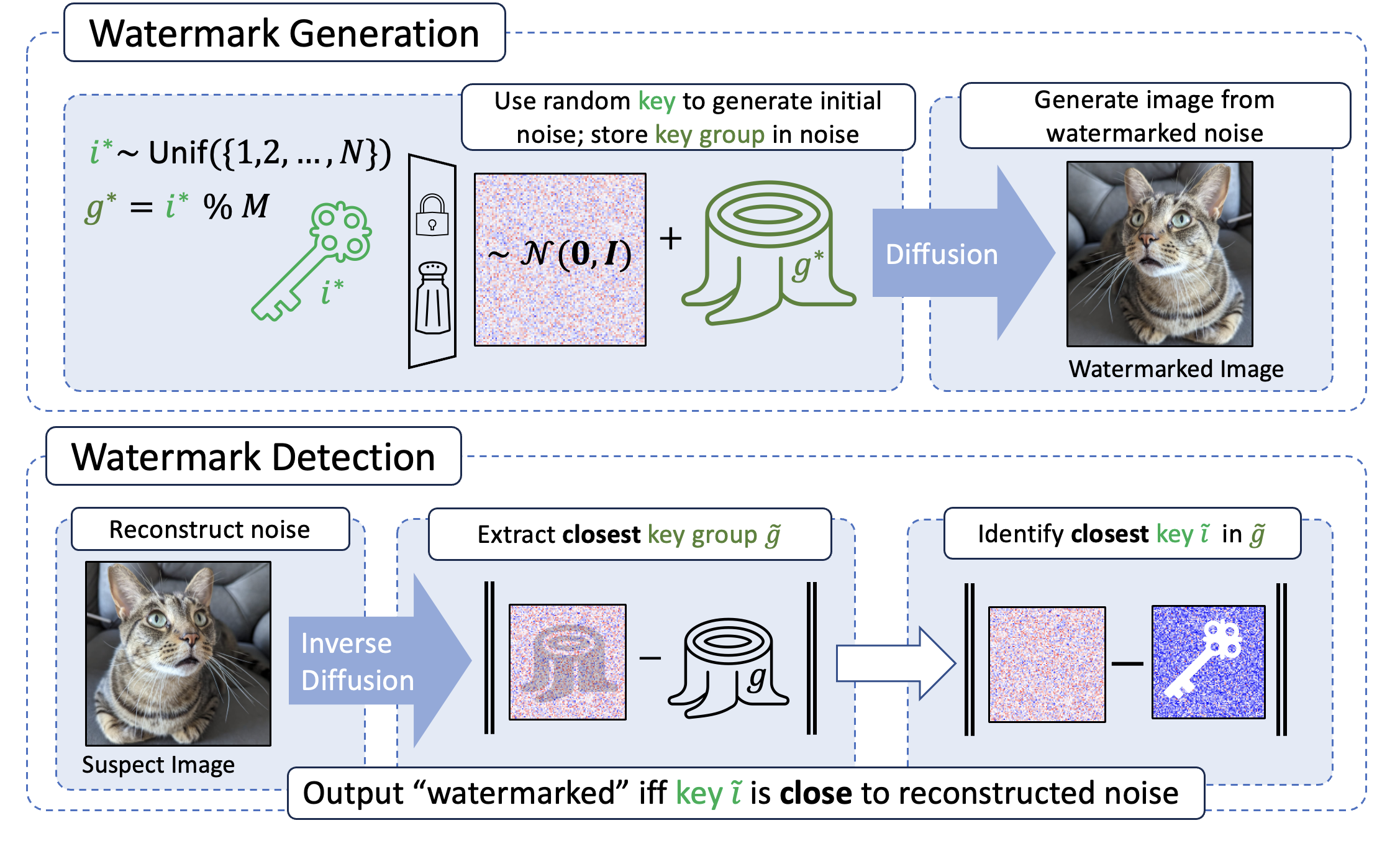}
    \caption{ \textbf{Illustration of the WIND Method for Robust Image Watermarking}.  The method is designed to use $N$ possible initial noises partitioned into $M$ groups. \textbf{Generation:} Using a secret salt and an index $i^*$, we securely and reproducibly generate initial noise $\mathbf{z}_{i^*}$. We then embed a group index $g^*$ of that noise to make easier retrieval possible using a Fourier pattern.
    Finally, we run diffusion with the embedded latent noise to produce a watermarked image.
    \textbf{Detection:} We reconstruct the initial noise $\tilde{\mathbf{z}}$.
    Next, we search over the possible group indices $g$ for the closest Fourier pattern to the one embedded in $\tilde{\mathbf{z}}$.
    We then look over initial noises in group $\tilde{g}$ to find the match.}
    \label{fig:graphic}
\end{figure}

Yet, given a sufficiently small set of initial noises (denoted as $N$) and an enormous number of images generated by a model, an attacker could potentially still collect many images sharing the same initial noise in order to perform removal and forgery attacks as was applied to previous methods \citep{yang2024steganalysisdigitalwatermarkingdefense}. Using many initial noises (a large value of $N$) will make such attacks much more difficult, if not infeasible. {Surprisingly, we find that a very large number of random initial noises remain distinguishable from one another, even after reconstructing the noise from a generated image.} However, a large value of $N$ might incur a negative effect on the runtime of our approach. 
In order to lower the effective quantity of noises we need to scan at detection while retaining strong robustness, we propose a two-stage efficient watermarking framework. We supplement our $N$ initial noise samples with $M$ Fourier patterns used as a \textit{group identifiers} - unique identifiers of the subset of initial noises we might have used for generating a given image (\Cref{fig:graphic}). During detection, we may first recover the group identifier (stage 1) and use it to find an exact match (stage 2).
Thus, we reduce our search space to the number of initial noises per group ($N / M$).



Our key contributions are as follows:
\begin{center}
  \begin{enumerate}
    \item We demonstrate that the initial noise used in the diffusion process is itself a distortion-free watermarking method for images (\Cref{sec:distortion-free}).
    \vspace{-0.1cm}
    \item We present \methodname{}, our two-stage method for effectively using the initial noise as a watermark (\Cref{sec:method}). 
    \vspace{-0.1cm}
    \item We demonstrate that \methodname{} achieves state-of-the-art results for its robustness to removal and forgery attempts (\Cref{sec:experiments}). 
  \end{enumerate}
\end{center}


\begin{figure}[t]
    \centering
    \includegraphics[width=0.7\textwidth]{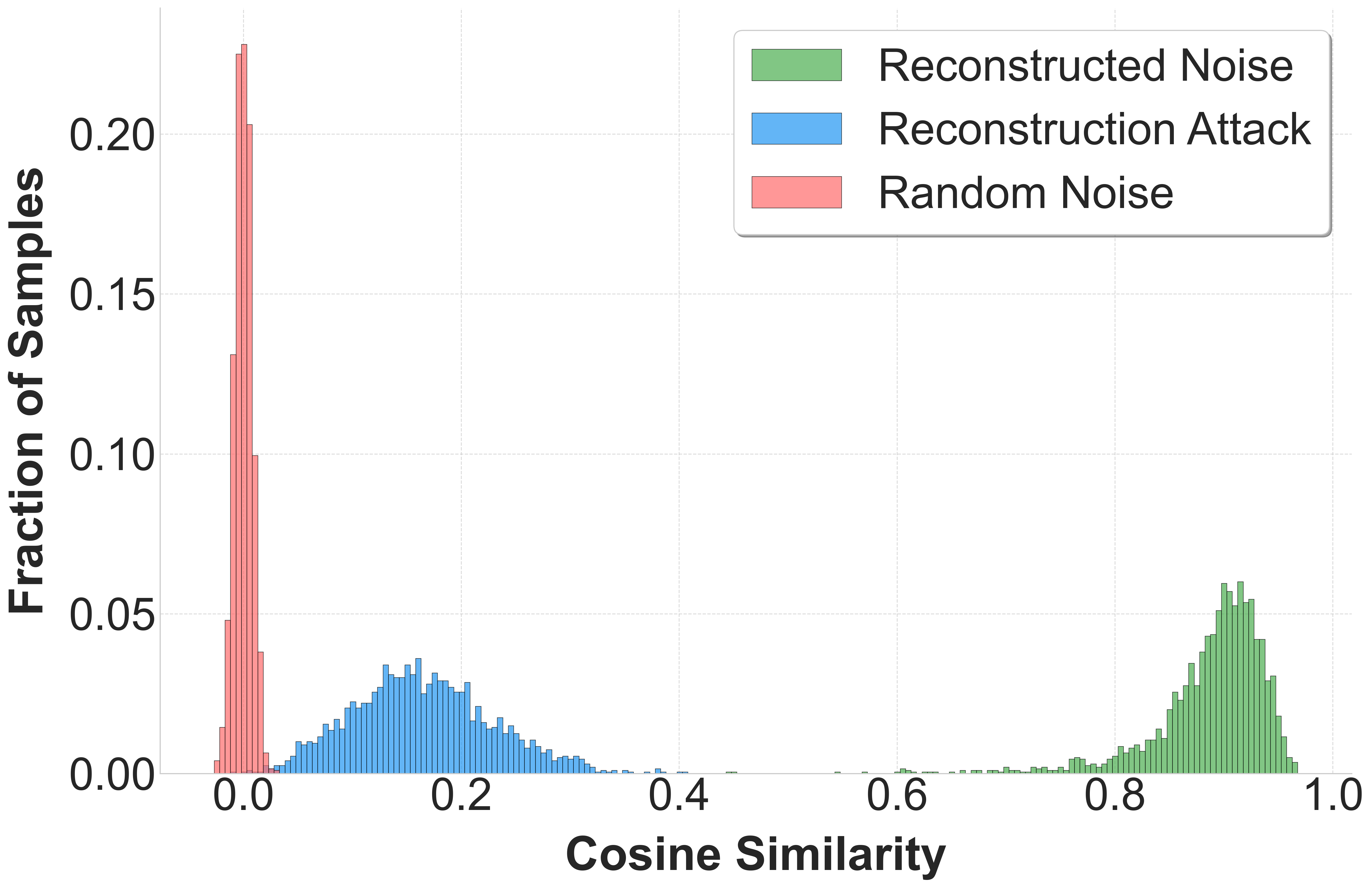}
    \caption{Cosine similarity distribution between initial noise, and: (i) a noise reconstructed from a watermarked image generated with the same noise (reconstructed noise) (ii) a noise reconstructed from a forged image using a public model to imitate our watermarked image  (reconstruction attack, described in \Cref{sec:distortion-free}). (iii) Random noise. These results are reliant on the approximate inversion of DDIM without the ground-truth prompt.}
    \label{fig:true_random}
\end{figure}

\section{Preliminaries}
\label{sec:preliminaries}

\subsection{Threat Model}

In a watermarking scheme we usually consider the owner, trying to mark images as an output of their model; and an attacker, trying to remove or forge the watermark on unrelated images.

\textbf{The Owner} releases a private model (diffusion model in our case) that clients can access through an API, allowing them to generate images that contain a watermark. The watermark is designed to have a negligible impact on the quality of the generated images. 
There are a few settings regarding the watermark detection, including public information and private information watermarking \citep{cox2007digital, wong2001secret}. We focus on the setting where the watermark is detectable only by the owner, enabling them to verify whether a given image was generated by their model using private information.

\textbf{The Attacker} uses the API to generate one image or more and subsequently attempts to launch a malicious attack aimed at either removing the watermark or forging the embedded watermark into unrelated images, with the intention of using the image or watermark for unauthorized purposes.

\subsection{Diffusion Models Inversion} \label{inv}

Diffusion model inversion aims to find the reconstructed noise representation of a given data point, effectively reversing the generative process.
Let $T$ be the number of diffusion steps in both the generation and inversion processes.
In the standard generation process, we start with noise $\tilde{\mathbf{x}}_T$ drawn from an appropriately scaled Gaussian and iteratively apply $\tilde{\mathbf{x}}_{t} = \tilde{\mathbf{x}}_{t+1} +  \mathbf{\epsilon}_{\Theta}(\tilde{\mathbf{x}}_{t+1})$, where $\mathbf{\epsilon}_{\Theta}$ is a trained model that predicts the noise to be removed and $t \in [T]$ is the time step describing how much noise should be removed in each stage. 
Conversely, the inversion process begins with a data point $\hat{\mathbf{x}}_0$ and moves towards its reconstructed noise representation by applying $\hat{\mathbf{x}}_{t+1} = \hat{\mathbf{x}}_{t} - \mathbf{\epsilon}_{\Theta}(\hat{\mathbf{x}}_{t})$. This process relies on the assumption that $\epsilon_{\Theta}(\hat{\mathbf{x}}_{t+1}) \approx \epsilon_{\Theta}(\hat{\mathbf{x}}_{t})$, allowing us to approximately invert the diffusion process by adding the predicted noise \citep{ho2020denoising,song2022denoisingdiffusionimplicitmodels}. {DDIM's efficient sampling allows this technique to be particularly useful. \mbox{\citep{song2022denoisingdiffusionimplicitmodels}}.}

\vspace{0.1cm}
\subsection{Tree-Ring and RingID Watermarks}

In order to watermark images in a human-imperceptible and robust way, previous works have encoded specific patterns in the Fourier space of the initial noise.
\textit{Tree-Ring} \citep{wen2023tree} first transforms the initial noise into the Fourier space. A key pattern is then embedded into the center of the transformed noise. The noise is subsequently transformed back into the spatial domain. 
During the detection phase, the diffusion process is inverted, and the Fourier domain is examined to verify the presence of the imprinted pattern. 
\textit{RingID} \citep{ci2024ringid} shows that Tree-Ring struggles to distinguish between different keys. Therefore, the number of unique keys (distinguishable from one another) that can be embedded with Tree-Ring is low. They increase the possible number of unique keys that can be encoded by modifying the patterns embedded in the  Fourier domain.


    

\noindent\textbf{Systematic Distribution Shifts in Generated Images Enable Attacks.} 
\label{pre:avg_att}
Systematic distribution shifts in the generated content make it easier to verify the existence of a watermark. However, in the case of Tree-Ring and other watermarking techniques, it also opens up an avenue of attack \citep{wen2023tree, yang2024gaussian, xian2024raw, bui2023rostealsrobuststeganographyusing}.  Emblematic is the method of \cite{yang2024steganalysisdigitalwatermarkingdefense}, whose attack approximates the difference between watermarked and non-watermarked images by estimating the difference between their averages in pixel space. Increasing the number of images with the watermark can improve the accuracy of the approximation. Removal attacks typically rely on paired images, but the forgery attack remains effective even when watermarked and non-watermarked images are unpaired. 




\section{Initial Noise is a Distortion-Free Watermark}
\label{sec:distortion-free}

Watermarks that systematically perturb the distribution of image generations are vulnerable to removal and forgery attacks. 
A distortion-free watermarking method, by contrast, may be more robust~\citep{yang2024steganalysisdigitalwatermarkingdefense}.
Our first finding is that the initial noise already in standard use in diffusion models can be such a watermark.

Let $N$ be the number of initial noises we can generate.
We will secure our watermarking process with a long, secret salt $s$.
We begin by sampling a psuedo-random (reproducible) initial noise.
Let $i^* \sim \textrm{Unif}([N])$ be the index of the initial noise.
We will use a hash function to get a seed $\textrm{hash}(i^*, s)$.
Plugging the seed into a pseudo-random generator, we generate a reproducible initial noise vector $\mathbf{z}_{i^*} \sim \mathcal{N}(\mathbf{0}, \mathbf{I})$ drawn from a centered Gaussian distribution.
When we generate fewer than $N$ images, we can use each initial noise at most once, and the noise appears distortion-free.
We discuss the case when the number of images exceeds $N$ in \Cref{app:distortion_free}.



\begin{algorithm}[h!]
\caption{Generation Algorithm}
\begin{algorithmic}[1]
    \State \textbf{Input:} $N$: number of initial noises, $M$: number of groups, $s$: secret salt, $p$: prompt, $\Theta$: private model weights
    \State Sample initial noise index $i^* \sim \textrm{Unif}([N])$
    \State Compute group identifier $g^* = i^* \% M$
    \Comment{Modulus of initial noise index}
    \State Calculate embedding of the group identifier $g_{emb}(g^*)$
    \State Securely generate \texttt{seed} $=$ \texttt{hash}$(i^*, s)$ \Comment{Apply cryptographic hash function}
    \State Sample $\mathbf{z}_{i^*} \sim \mathcal{N}(\mathbf{0}, \mathbf{I})$ from a pseudorandom generator with \texttt{seed}
    \State Add the identifier embedding $g_{emb}(g^*)$ to $\mathbf{z}_{i^*}$ to get $\mathbf{z}_{i^*\_emb}$
    \State \textbf{return} \texttt{image} $= G_{\Theta}(\mathbf{z}_{i^*\_emb}, p)$ \Comment{Diffusion process $G$ with weights $\Theta$}
\end{algorithmic}
\label{alg:gen}
\end{algorithm}
\begin{algorithm}[h!]
\caption{Detection Algorithm \textit{\textbf{(WIND\_fast)}}}
\begin{algorithmic}[1]
    \State \textbf{Input:} \texttt{image}: (possibly) watermarked image, $N$: number of initial noises, $M$: number of groups, $s$: secret salt, $\Theta$: private model weights, $\tau:$ threshold for detection
    \State Recover reconstructed noise $\tilde{\mathbf{z}} = G^{-1}_\Theta($\texttt{image}$)$ \Comment{Inverse diffusion with private weights}
    \State Extract closest group identifier $\tilde{g}$ from group identifier embedding in $\tilde{\mathbf{z}}$
    \For{$i \in [N]$ such that $i \% M = \tilde{g}$} \Comment{Search over subset of initial noise indices}
        \State Build initial noise $\mathbf{z}_i$ using secret salt $s$ and \texttt{hash} \Comment{As in \Cref{alg:gen}}
        \State Compare $\mathbf{z}_i$ to $\tilde{\mathbf{z}}$ after removing Fourier embedding $\tilde{g}$
    \EndFor
    \If{any noises are closer than threshold $\tau$}
    \State Declare ``watermarked''
    \Else
    \State Declare ``not watermarked''
    \EndIf
\end{algorithmic}
\label{alg:det_fast}
\end{algorithm}

\noindent\textbf{Empirical validation of initial noise watermarking.}
To empirically validate our claim that the initial noise can serve as a watermark, we will compute the cosine similarity between the initial noise $\mathbf{z}_{i^*}$ and (i) random noise $\mathbf{z} \sim \mathcal{N}(\mathbf{0}, \mathbf{I})$, (ii) reconstructed noise $\tilde{\mathbf{z}}$ when we have access to the private model weights, and (iii) the reconstructed noise $\tilde{\mathbf{z}}^\textrm{attack}$ from an image imitating our noise pattern without access to the private model weights (we used another checkpoint of Stable Diffusion-v2 \citep{rombach2022highresolutionimagesynthesislatent}, as it is the most similar model to the Stable Diffusion 2.1 model we use). 
We provide the results in \Cref{fig:true_random} and \Cref{tab:inversion_att} and find that the pattern reconstructed with the private model is indeed significantly more correlated with the initial noise compared to both a random pattern and a pattern reconstructed with another model. We analyze the resulting probabilistic guarantee under \textit{reconstruction attack} below.

\textbf{Watermarking Process.} During the watermarking process, we create an image \texttt{image} through diffusion with the private model weights $\Theta$ conditioned on a private text prompt $p$.
Formally, \texttt{image} $= G_{\Theta}(\mathbf{z}_{i^*}, p)$.
We obtain the reconstructed noise that we use for detection via an inverse diffusion process $G^{-1}$.
Formally, $\tilde{\mathbf{z}} = G^{-1}_\Theta($\texttt{image}$)$. If the cosine similarity $\sim(\mathbf{z}_{i^*}, \tilde{\mathbf{z}})$ is below a threshold $\tau$, we declare the image watermarked.

\textbf{Reconstruction Attack.} An attacker trying to forge the watermarked image will not have access to our private weights, instead, they will have some other weights $\Theta'$ (\cite{keles2023finegrainedhardnessinvertinggenerative} demonstrates that inverting a generative model is a challenging task).
Yet, they may attempt to recover the initial noise using a public watermarked image and a public model.
Let $\tilde{\mathbf{z}}' = G^{-1}_{\Theta'}($\texttt{image}$)$.
Then, with this initial noise, they will generate a forged image with (possibly offensive) text prompt $p'$, producing \texttt{image}' $=G_{\Theta'}(\tilde{\mathbf{z}}', p')$.
Finally, the model owner will attempt to detect whether the forged image is watermarked by applying the inverse diffusion process with the private model weights to the forged image.
Let $\tilde{\mathbf{z}}^{\textrm{attack}} = G^{-1}_\Theta($\texttt{image}'$)$. As an upper bound on the capability of this attack, we perform the adversarial generation with the same prompt.

Strikingly, we find that the similarity between the true noise and the noise reconstructed with the model weights is almost always greater than a relatively large threshold $\tau = 0.5$ ($p$ value $<10^{-3}$, \Cref{fig:true_random}).
At the same time, the reconstructed similarity from the image made by an attacker using the reconstruction attack \texttt{sim}$(\mathbf{z}_{i^*}, \tilde{\mathbf{z}}^\textrm{attack})$, along with the similarity to random vectors \texttt{sim}$(\mathbf{z}_{i^*}, \mathbf{z})$ are both much smaller. Namely, they are respectively $z=5.3$ and $z=9.4$ standard deviations below the threshold (\Cref{tab:inversion_att}).
Taken together, these results mean that the probability $p$ of a non-watermarked image mistakenly labeled as watermarked is very low in both cases. For a random unrelated noise, the probability to confuse it is as the initial noise is $p < e^{(\frac{\tau ^ 2}{2\sigma ^ 2})} < 10^{-19}$, allowing for practically perfect detection of the correct pattern among a very large number of initial noise instances. 

\noindent\textbf{Runtime considerations.} 
Our method requires searching over all $N$ watermarks, leading to a naive runtime complexity of \(\mathcal{O}(N)\). However, more efficient algorithms for similarity-based search, such as HNSW~\citep{malkov2018efficientrobustapproximatenearest}, can reduce this complexity to \(\mathcal{O}(\log N)\), at the expense of additional memory usage. For large enough values of $N$, this cost may eventually become undesirable. Along with our goal of maintaining high robustness as the number of keys increases, this motivates a more efficient method, which we present in the next section.


\section{Method}
\label{sec:method}
\subsection{WIND: Two-stage Efficient Watermarking} \label{framework}
While always using the same initial noise for our model might imply good robustness properties against removal, to make forgery and removal more difficult, it is generally preferable to maintain a large set of $N$ initial noises to be used by the model. Moreover, using a large number of different noises $N$ may serve as different keys, encoding some metadata about each image. This metadata might include information about the specific model that generated it, as well as additional information about the generation for further validation of the image source, once detected.

\begin{figure}[t!]
    \centering
    \begin{subfigure}[b]{0.45\textwidth}
        \centering
        \includegraphics[width=\textwidth]{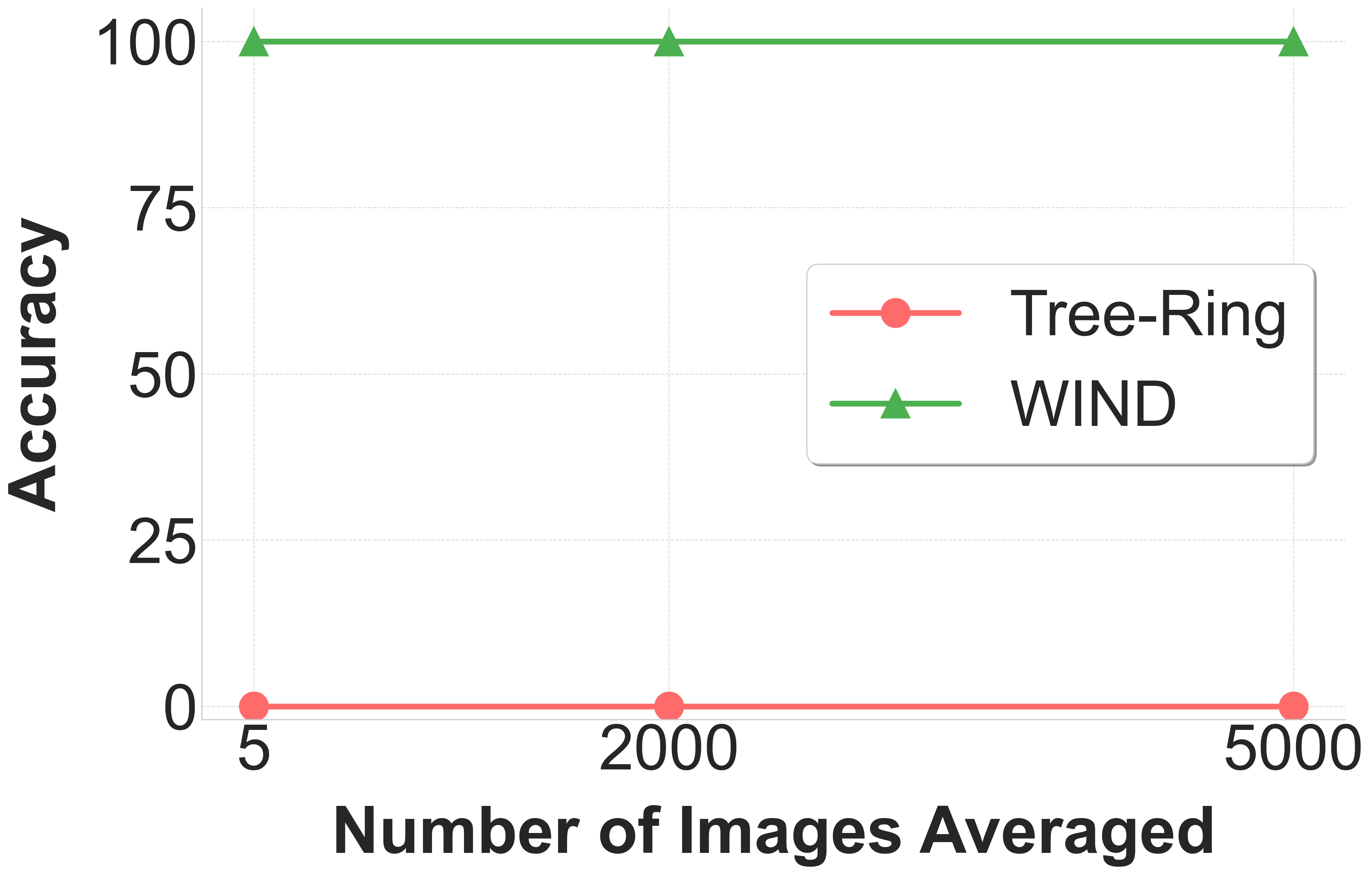}
        \caption{Forgery Performance}
        \label{fig:forgery_performance}
    \end{subfigure}
    \begin{subfigure}[b]{0.45\textwidth}
        \centering
        \includegraphics[width=\textwidth]{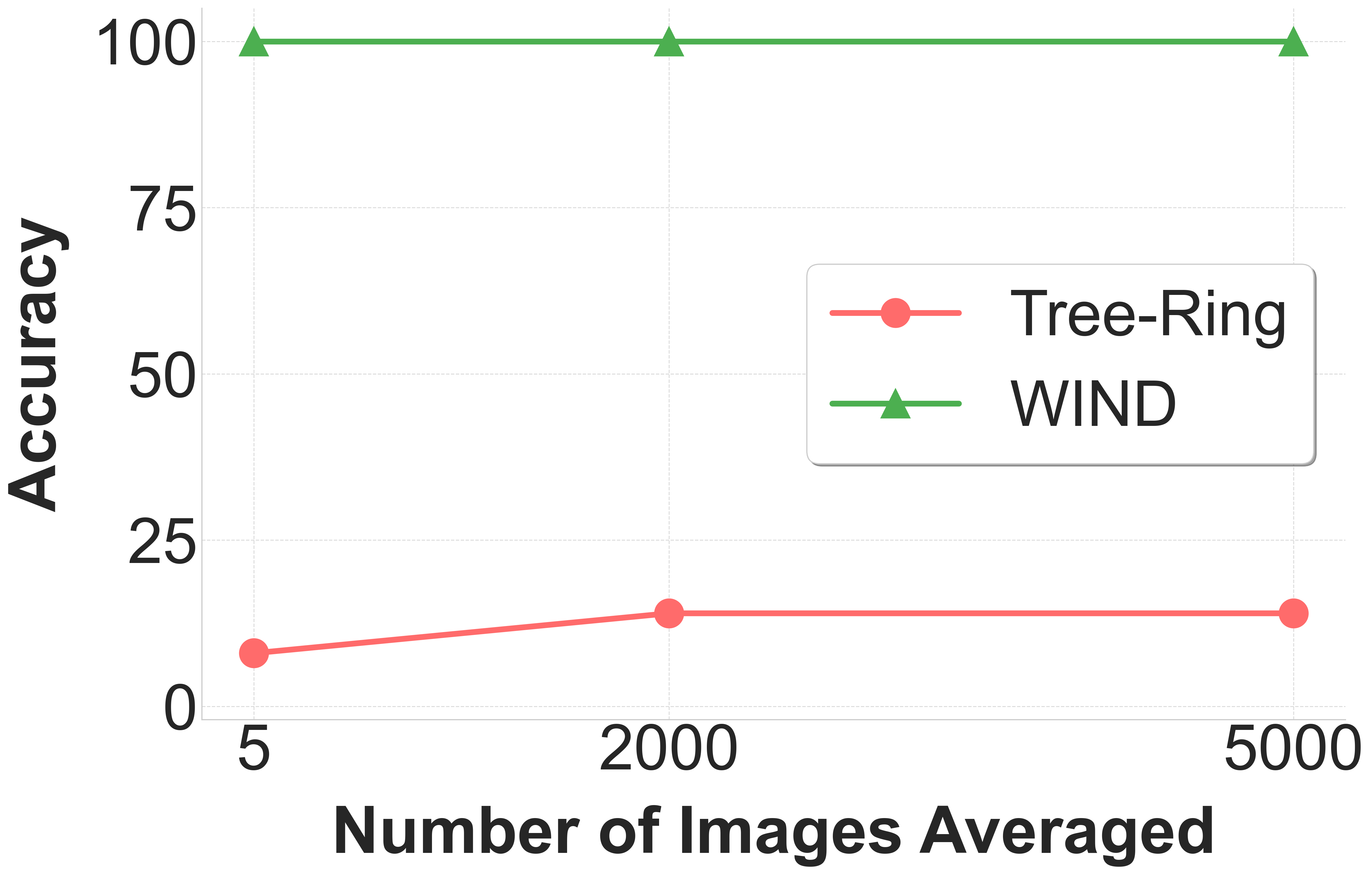}
        \caption{Removal Performance}
        \label{fig:removal_performance}
    \end{subfigure}
    \caption{Detection accuracy for forgery and removal attacks using \cite{yang2024steganalysisdigitalwatermarkingdefense}. A value of $0\%$ represents a watermark failure (the attacker successfully removed the watermark or forged it onto another image), while $100\%$ indicates a perfect defense (no watermark removal or forgery occurred).}
    \label{fig:avg_att}
\end{figure}

In order to make the search over a large number of noises more efficient, we introduce a two-stage efficient watermarking approach we name \textit{\textbf{WIND}} (\textbf{{W}}atermarking with \textbf{{I}}ndistinguishable and Robust \textbf{{N}}oise for \textbf{{D}}iffusion Models). First, we initialize $M$ groups of initial noise, each group associated with its own Fourier-pattern key. 
In contrast to prior work, we employ these Fourier patterns not as a watermark, but as a \textit{group identifier} to reduce the search space.

\begin{wrapfigure}[26]{r}{0.5\textwidth}
    \centering
    \includegraphics[width=\linewidth]{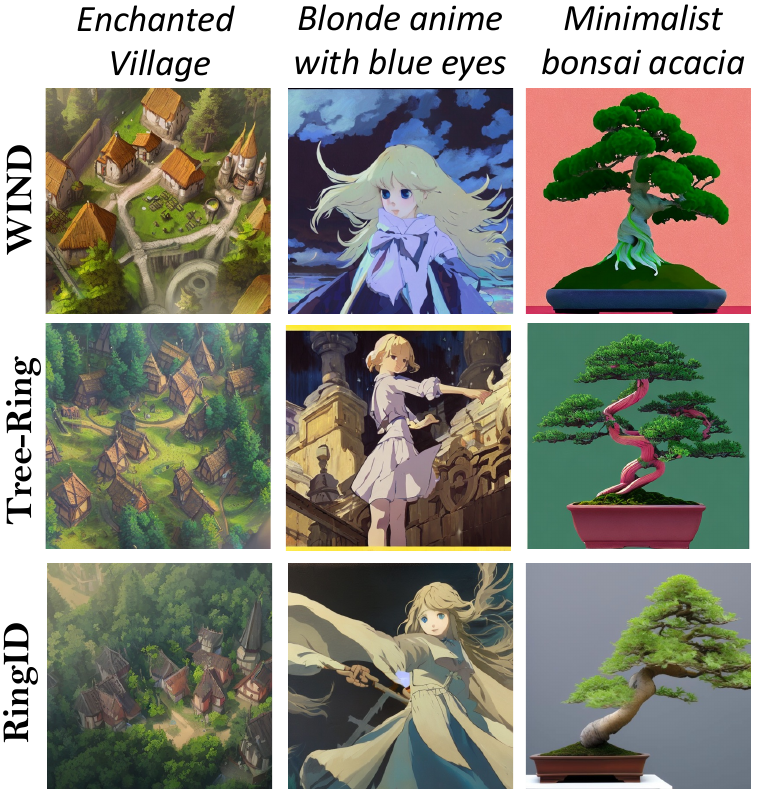}
    \caption{Qualitative results of watermarked images generated using WIND, Tree-Ring, and RingID. {See \mbox{\Cref{app:additional_results}} for quantitative results. See \mbox{\Cref{appendix:generated_images}} for additional qualitative results.}}
    \label{fig:comp}
\end{wrapfigure}

For each image generation, we randomly select an index for the initial noise, denoted as $i^* \in [N]$. We use a group identifier $g^* = i^* \% M$, where $\%$ denotes the modulus operation.
We embed $g^*$ in the Fourier space of the latent noise (similar to \cite{wen2023tree}).
During detection, we reconstruct the latent noise and find the group identifier $\tilde{g}$ that is closest to the Fourier pattern embedded in the image. We account for possible rotations and crops in our search (see \Cref{sec:implementation-details}).
We then search over all indices $i$ such that $\tilde{g} = i \% M$.
In this way, the search space has size $N/M$ rather than $N$. We include an algorithm box for generation (\Cref{alg:gen}) and detection (\Cref{alg:det_fast}).

In the following part, we refer to two variants of our method: (i) \textit{\textbf{WIND$_{\text{fast}}$}}  where we assume the used initial noise belongs to the identified group $\tilde{g}$ and check similarity only to noise patterns in this group. (ii)
\textit{\textbf{WIND$_{\text{full}}$}} where we check all $N$ possible initial noises if we can't find a match within the detected group (the gap between the similarity of the correct noise and random noises, as shown in \mbox{\Cref{fig:true_random}}, allows us to still determine whether the correct noise has been identified). The \textit{\textbf{WIND$_{\text{full}}$}} method is slower but more robust to removal attacks that might interfere with the Fourier pattern. {Empirical validation can be found in \mbox{\Cref{sec:experiments}.}
Additional ablations and results can be found in \mbox{\Cref{app:additional_results}.}} We provide empirical runtime analysis of our method in \Cref{ab:emp}.
\subsection{Resilience to Forgery}\looseness=-1
{In addition to empirical evaluations of specific attacks as in \mbox{\Cref{fig:true_random,fig:avg_att}}; we discuss below the attacker's ability to infer knowledge about the used noise pattern across different watermarked images.}
Even if the attacker is able to obtain information about a specific initial noise $\mathbf{z}_i$ for an index $i$, the other noise vectors for $j \neq i$ are still safe\footnote{We note that obtaining a single noise pattern might not be enough to effectively forge the watermark, as the model owner may encode this pattern with additional metadata as described in \Cref{framework}}.
This is because we use a cryptographic hash function and a secret salt.
Formally, \Cref{thm:crypto_main} shows that, as long as the cryptographic hash function remains unbroken and the secret salt is kept private, the watermarking algorithm maintains its security properties against even very powerful adversaries.  

\begin{restatable}{theorem}{cryptothm}[Cryptographic Security]\label{thm:crypto_main}
Let \textnormal{\texttt{hash}}$: {0,1}^* \rightarrow {0,1}^\ell$ be an unbroken cryptographic hash function used in our watermarking algorithm, with inputs $i^* \in [N]$ and a secret salt $s$. Assume $s$ is sufficiently long and randomly generated. Then, even if an adversary obtains: the group number $g^*$, the initial noise index $i^*$, the initial noise $\mathbf{z}_{i^*}$, and even the corresponding output of the hash function \textnormal{\texttt{seed}}, the adversary cannot:
\begin{enumerate}
    \item Recover the secret salt $s$,
    \item Generate valid reconstructed noise $\mathbf{z}_j$ for any other initial noise index $j \neq i$
\end{enumerate}
\end{restatable}

We defer the proof to \Cref{app:crypto_thm}.

\begin{table}[t]
    \centering
    \caption{Comparison of correct watermark detection accuracy between WIND and previous image watermarking approaches under various image transformation attacks. $\text{WIND}_M$ denotes the use of $M$ groups, with the total number of noises ($N$) specified in the \enquote{Keys} column. {A broader comparison with additional methods can be found in \mbox{\Cref{tab:ab_performance}}.}}
    \begin{tabular}{@{}llccccccccc@{}}
        \toprule
        \textbf{Method} & \textbf{Keys} & \textbf{Clean} & \textbf{Rotate} & \textbf{JPEG} & \textbf{C\&S} & \textbf{Blur} & \textbf{Noise} & \textbf{Bright} & \textbf{Avg $\uparrow$} \\ 
        \midrule
        \multirow{3}{*}{Tree-Ring} 
            &  32    & 0.790 & 0.020 & 0.420 & 0.040 & 0.610 & 0.530 & 0.420 & 0.404 \\ 
            &  128   & 0.450 & 0.010 & 0.120 & 0.020 & 0.280 & 0.230 & 0.170 & 0.183 \\ 
            &  2048  & 0.200 & 0.000 & 0.040 & 0.000 & 0.090 & 0.070 & 0.060 & 0.066 \\ 
        \midrule
        \multirow{3}{*}{RingID} 
            &  32      & \textbf{1.000} & \textbf{1.000} & \textbf{1.000} & 0.530 & 0.990 & \textbf{1.000} & 0.960 & 0.926 \\ 
            & 128     & \textbf{1.000} & 0.980 & \textbf{1.000} & 0.280 & 0.980 & \textbf{1.000} & 0.940 & 0.883 \\ 
            &  2048    & \textbf{1.000} & 0.860 & \textbf{1.000} & 0.080 & 0.970 & 0.950 & 0.870 & 0.819 \\ 
        \midrule
        \multirow{1}{*}{WIND$_{\text{fast}_{128}}$} 
            & 100000 & \textbf{1.000} & 0.780 & \textbf{1.000} & 0.470 & \textbf{1.000} & \textbf{1.000} & 0.960 & 0.887 \\ 
            
        WIND$_{\text{fast}_{2048}}$ & 100000& \textbf{1.000} &0.870&0.960&0.060&0.960&0.950&0.900&0.814 \\ 

        \midrule
        \multirow{1}{*}{WIND$_{\text{full}_{128}}$} 
            &  100000 & \textbf{1.000} & 0.780 & \textbf{1.000} & 0.850 & \textbf{1.000} & \textbf{1.000} & \textbf{1.000} & \underline{0.947} \\ 
        WIND$_{\text{full}_{2048}}$ & 100000& \textbf{1.000} &0.880&\textbf{1.000}&\textbf{0.930}&\textbf{1.000}&0.990&0.980 &\textbf{0.969} \\ 
        \bottomrule
    \end{tabular}
    \label{tab:performance}
\end{table}

\noindent\subsection{Watermarking Non-Synthetic Images.} 
\label{subsec:non-synthetic}
By using diffusion inpainting, our watermark can be applied to a natural image. Later, by inverting the inpainted image, we can verify the presence of the watermark. 

\begin{figure}[t!]
    \centering
    \begin{tabular}{cccc}
        \textbf{Before} & \textbf{After} & \textbf{Before} & \textbf{After} \\[6pt]
        
        \includegraphics[width=0.2\textwidth]{images/fig_inp_main/org1.png} &
        \includegraphics[width=0.2\textwidth]{images/fig_inp_main/wm1.png} &
        \includegraphics[width=0.2\textwidth]{images/fig_inp_main/org2.png} &
        \includegraphics[width=0.2\textwidth]{images/fig_inp_main/wm2.png} \\[12pt]
        
        
        \includegraphics[width=0.2\textwidth]{images/fig_inp_main/org3.png} &
        \includegraphics[width=0.2\textwidth]{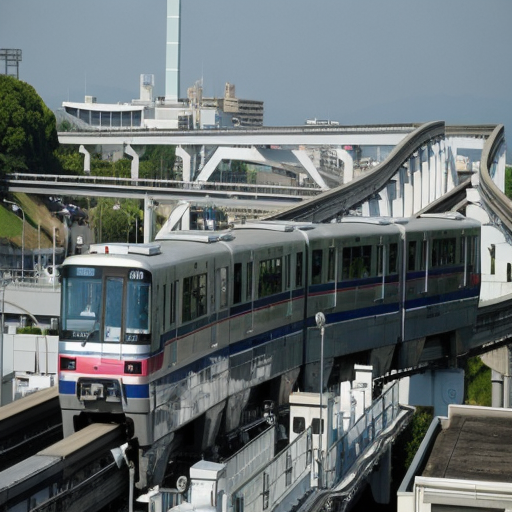} &
        \includegraphics[width=0.2\textwidth]{images/fig_inp_main/org4.png} &
        \includegraphics[width=0.2\textwidth]{images/fig_inp_main/wm4.png} \\
    \end{tabular}
    \caption{Comparison of COCO images before and after watermarking via inpainting.}
    \label{fig:inp}

\end{figure}

As demonstrated in \Cref{fig:inp}, our inpainting method injects a watermark with minimal visual impact, preserving the original image's integrity. See \mbox{\Cref{app:additional_results}} for additional results.

\section{Experiments}
\label{sec:experiments}


\subsection{Watermark Robustness}
\label{exp:rob}

\noindent\textbf{Setting.}
For a fair comparison with previous methods \citep{ci2024ringid,wen2023tree}, we employed Stable Diffusion-v2 \citep{rombach2022highresolutionimagesynthesislatent}, with 50 inference steps for both generation and inversion. Other implementation-details can be found in \Cref{sec:implementation-details}.

\noindent\textbf{Image Transformation Attacks.} Following previous methods \citep{wen2023tree, ci2024ringid} we applied these image transformations to the generated images:  $75^\circ$ rotation, $25\%$ JPEG compression, $75\%$ random cropping and scaling (C \& S), Gaussian blur with an $8 \times 8$ filter size, Gaussian noise with $\sigma = 0.1$, and color jitter with a brightness factor uniformly sampled between 0 and 6.
  In \Cref{tab:performance} we compare our methods to both Tree-Ring and RingID. As the results demonstrate, using multiple keys with RingID \citep{ci2024ringid} is possible. Yet, it remains vulnerable to cropping and scaling attacks. In contrast, WIND effectively addresses this challenge. It enables accurate watermark detection under all image transformation attacks. 


\noindent\textbf{Steganalysis Attack.} We assess the robustness of our method against the attack proposed by \cite{yang2024steganalysisdigitalwatermarkingdefense}, which is capable of forging and removing the Tree-Ring and RingID keys. As discussed in \Cref{pre:avg_att}, this attack attempts to approximate the watermark by subtracting watermarked images from non-watermarked images.
The results, presented in \Cref{fig:avg_att}, indicate that the attack can be used to forge or remove our RingID group identifier. Yet, it is unable to forge or remove our full watermark (initial noises). Even when the Fourier pattern type key is removed, our method remains robust in identifying the correct initial noise through an exhaustive search.

\begin{table}[t]
    \centering
    \begin{minipage}{0.4\textwidth}
        \centering
    \centering
    \caption{Cosine similarity between the initial noise and the inversed noise before and after the regeneration attack. {Also see \mbox{\Cref{app:additional_results}.}}}
    \begin{tabular}{@{}lcc@{}}
        \toprule
        \textbf{Condition} & \textbf{Mean} & \textbf{STD} \\
        \midrule
        Original Image  & 0.888 & 0.053 \\
        Attacked Image & 0.824 & 0.062 \\
        Unrelated Image & 0.000 & 0.008 \\
        \bottomrule
    \end{tabular}
    \label{tab:reg_att}
    \end{minipage}
    \hspace{0.3cm}
    \begin{minipage}{0.4\textwidth}
    \centering
    \caption{FID scores of WIND compared to previous watermarking approaches.}
    \begin{tabular}{@{}lcc@{}}
        \toprule
        \textbf{Method} & \textbf{FID $\downarrow$} \\ \midrule
        DwtDctSvd  &  25.01 \\
        RivaGAN  &  \underline{24.51} \\
        Tree-Ring  &  25.93 \\ 
        RingID  &  26.13 \\ 
        WIND  & \textbf{24.33} \\ 
        \bottomrule
    \end{tabular}
    \label{tab:fid}
    \end{minipage}
\end{table}

\begin{table}[t!]
    \centering
    \looseness=-1\caption{Cosine similarity between the initial noise used for generation and the inversed noise obtained through three inversion approaches. ``Private'' refers to models owner's model, while ``Public'' denotes external model.}
    \begin{tabular}{@{}lcc@{}}
        \toprule
        \textbf{Approach} & \textbf{Mean} & \textbf{Std} \\ \midrule \vspace{0.1cm}
        Gen (private) $\rightarrow$ Rev (private) & 0.888 & 0.053
        \\  \vspace{0.1cm}
        Gen (private) $\rightarrow$ Rev (public) $\rightarrow$ Gen (public) $\rightarrow$ Rev (private) & 0.166 & 0.063 
        \\  
        Random Noise & 0.000 & 0.053 \\  
        \bottomrule
    \end{tabular}
    \label{tab:inversion_att}
\end{table}

\noindent\textbf{Regeneration Attacks.} Recently, \cite{zhao2023invisibleimagewatermarksprovably} introduced a two-stage regeneration attack: (i) adding noise to the representation of a watermarked image, and (ii) reconstructing the image from this noisy representation. To assess the resilience of our approach to regeneration attacks, we applied the attack from \cite{zhao2023invisibleimagewatermarksprovably} to watermarked images generated by our model. As shown in \Cref{tab:reg_att}, the attack has a minimal impact on the distribution of the cosine similarities between the initial noise and the inverted noise. The attacked noise similarity still maintains a significant gap compared to random noise.

\noindent\textbf{Image Quality.}
To examine the performance of our inpainting method, we report the Fréchet Inception Distance (FID) \citep{heusel2018ganstrainedtimescaleupdate} on the MS-COCO-2017 \citep{lin2015microsoftcococommonobjects} training dataset in \Cref{tab:fid}. Notably, our method achieves the lowest FID among the compared methods, indicating a closer alignment with real images. While the image quality of our generated images can be understood analytically - no distortion for the initial noise methods (\Cref{sec:distortion-free}) and similar distortion to RingID for the two-stage method (\Cref{sec:method}) - we also include some images generated by our framework in \Cref{fig:comp}.

\section{Discussion and Limitations}
\label{sec:limitations}
\textbf{Editing a Given Image vs. Forging.} While forging our watermark by obtaining the initial noise is hard (\Cref{sec:distortion-free}), an easier path to obtaining harmful watermarked images might be to apply a slight edit to an already watermarked image. A harmful image in this context might include the addition of a copy-right infringing material, NSFW materials, or any other content the model owner wishes to avoid being associated with. Naturally, there is a trade-off between the severity of the applied edit and the edit's ability to preserve the initial watermark. We present one solution to mitigating this issue in the next discussion point.

\textbf{Storing a Database of Generations.} Model owners wishing to protect themselves from an attacker modifying a watermark image may keep a database of the past generations by their model. For these extreme cases, the model owner might only save the used prompts and initial noise seeds and use the reconstructed noise to retrieve the entire set of prompts used with that specific seed \citep{huang2024waterpool}. While this process may be resource-intensive, it is only required in the rare event that an attacker intentionally modifies a benign image into a harmful one while preserving the watermark.

\textbf{Private Model.}
Our watermark robustness is based to a large extent on the inability of an attacker to invert a model, which is empirically validated only for some attacks. Yet, as discussed in  \Cref{inv}, the ability to successfully invert our model may be nearly equivalent to the ability to steal the forward diffusion process, effectively stealing the model (in which case, any watermarking attempt might be deemed quite useless anyhow). Still, a better framing of the mathematical assumptions behind this claim is a limitation of this work, as an attacker might be able to learn something about the noise without full access to the model.

\textbf{Attacker's Advantage.} There exists a large set of diverse attacks aimed at watermark removal \citep{zhang2023watermarks,yang2024steganalysisdigitalwatermarkingdefense,zhao2023invisibleimagewatermarksprovably}, along with image transformations such as rotation and crops that also achieve some limited success against our watermark. As in many security applications, we suspect that an attacker capable enough will still be able to remove the watermark using new techniques we might not expect. However, a more robust watermark may nevertheless help to decrease the spread of false information. 

{Additional discussion and limitations can be found in \mbox{\Cref{app:additional_disc}}.}

\section{Conclusion}
In this work, we present a robust watermarking method that leverages the initial noises employed in diffusion models for image generation. By integrating existing techniques, we enhanced the approach to achieve improved efficiency while maintaining our robustness against various types of attacks. Furthermore, we outlined a strategy for applying our method to non-generated images through inpainting.

\bibliography{iclr2025_conference}
\bibliographystyle{iclr2025_conference}

\clearpage
\appendix
\appendixpage
\tableofcontents
\newpage




\section{Notation}
\label{app:notations}
\begin{table}[h]
\centering
\caption{Notations used in the paper.}
\renewcommand{\arraystretch}{1.5}  %
\begin{tabular}{ll}
\hline
$N$ & Number of initial noises \\
$M$ & Number groups  \\
$s$ & Secret salt for cryptographic security \\
$i$ & Index of initial noise: $i \in [N]$ \\
$g$ & Index of group: $g = i \% M$ \\
\texttt{hash} & A cryptographic hash function \\
$\mathbf{z}$ & Initial noise \\
$\tau$ & Threshold for declaring an image is watermarked \\
$T$ & Number of diffusion steps \\
$\Theta$ & Weights of a diffusion model \\
$p$ & Text prompt for diffusion \\
$G_\Theta$ & Diffusion model with weights $\Theta$ \\ 
$G^{-1}_\Theta$ & Inverse diffusion model with weights $\Theta$ \\
\hline
\end{tabular}
\end{table}

\section{Related Works}
\label{sec:related-works}

\textbf{Memorization in Diffusion Models.}
Diffusion models \citep{ho2020denoising,sohl2015deep} have demonstrated a capacity not only to generalize but also to memorize training data. This can lead to the reproduction of specific patterns or, in some cases, exact content from the training set, including sensitive or proprietary information. This memorization poses significant risks of unintended intellectual property leakage, particularly in large-scale generative models.
Several studies have shown that information from training data can be extracted from diffusion models \citep{carlini2023extracting, somepalli2023understandingmitigatingcopyingdiffusion, carlini2023extractingtrainingdatadiffusion, gu2023memorizationdiffusionmodels, somepalli2023understanding}. The risk of memorization in diffusion models underscores the need for accountability among model owners. 

\textbf{Image Watermarking.}
Image watermarking is essential for protecting intellectual property, verifying content authenticity, and maintaining the integrity of digital media. The field ranges from traditional signal processing techniques to recent deep learning methods \citep{potdar2005survey, singh2023comprehensive}.

Among Early watermarking strategies, one of the classical methods is Least Significant Bit (LSB) embedding, which modifies the least significant bits of image pixels to imperceptibly embed watermarks \citep{wolfgang1996watermark}. Another classical approach utilizes frequency-domain transformations and Singular Value Decomposition (SVD) to hide watermarks within image coefficients. \citep{chang2005svd, al2007combined}.

Recent developments leverage deep learning for watermarking. For instance, HiDDeN \citep{zhu2018hiddenhidingdatadeep} introduced an end-to-end trainable framework for data hiding. RivaGAN \citep{zhang2019robustinvisiblevideowatermarking} utilizes adversarial training to embed watermarks, while \cite{lukas2023ptwpivotaltuningwatermarking} proposed an embedding technique that optimizes efficiency by avoiding full generator retraining.

\textbf{Watermarking for Diffusion Models.} Existing watermark methods for diffusion models can be divided into three categories:

(i) Post-processing methods which adjust image features to embed watermarks \citep{zhao2023recipe, fernandez2023stablesignaturerootingwatermarks}. This approach alters the generated image distribution. However, recent work by \cite{zhao2023generative} shows that pixel-level perturbations are removable by regeneration attacks makes. To date, this approach is not robust.

(ii) Fine-tuning-based approaches combine the watermark within the generation process \citep{zhao2023recipe, xiong2023flexible, liu2023watermarking, fernandez2023stable, cui2023diffusionshield}. To date, these methods have robustness issues as well \citep{zhao2023invisibleimagewatermarksprovably}.

(iii) Tree-Ring introduced an approach to proposing a method to imprint a tree-ring pattern into the initial noise of a diffusion model \citep{wen2023tree}. Each pattern is used as a key, which is added in the Fourier space of the noise. The verification of the presence of the key involves recovering the initial noise from the generated image and checking if the key is still detectable in Fourier space. 
This approach makes Tree-Ring and its follow-up works the most robust approach against regenration attacks \citep{zhao2023invisibleimagewatermarksprovably, an2024wavesbenchmarkingrobustnessimage, gunn2024undetectablewatermarkgenerativeimage}.

Recently, \cite{yang2024steganalysisdigitalwatermarkingdefense} took advantage of the distribution shift present in Tree-Ring that occurs with impainting keys and arranged the first successful black box attack against it, as we detailed in \Cref{pre:avg_att}.

\section{\texorpdfstring{{Additional Discussion and Limitations}}{Additional Discussion and Limitations}}
\label{app:additional_disc}

\paragraph{{{Relation to Other Initial Noise Watermarking Methods.}}}
{The seminal work by \mbox{\cite{wen2023tree}} innovated the use of initial noise in DDIM for watermarking. Most related to our work, \mbox{\cite{yang2024gaussian}} also embeds a watermark in the initial noise already used by a DDIM diffusion model. Yet, while \mbox{\cite{yang2024gaussian}} proposes a watermark that is distortion-free for a single image, it is not distortion-free when examining sets of images; therefore it is vulnerable to attacks such as \mbox{\cite{yang2024steganalysisdigitalwatermarkingdefense}}. We aim to be robust to attacks even when 
many images are examined together.}

{There are additional technical differences between our approach and \mbox{\cite{yang2024gaussian}}. Most notably: (i) Our work also studies applying our watermark to non-synthetic (natural) images, or images coming from other generative models. (ii) While \mbox{\cite{yang2024gaussian}} design a function to embed specific bits into the initial noise, we take another approach. Namely, we view the entire initial noise (with generation and inversion) as a noisy channel. Inspired by \mbox{\cite{shannon1949communication}}, we use a random encoding of the watermark identities into the channel.}

\textbf{Computational Requirements.} 
As discussed in \Cref{sec:distortion-free}, our similarity search can be accelerated given well-known methods. Yet, the computational requirements of our method might be limiting when trying to use our method on edge devices. However, similarly to Tree-Ring and Ring-ID \citep{wen2023tree,ci2024ringid} our method assumes a private model, which is usually not deployed on edge devices anyhow.

\paragraph{{{Trade-Offs Between the Watermarking Overhead, and Detection Accuracy.}}}
{We suggest the following variants of our method for different possible requirements of runtime scaling, detection robustness, and ease of adaptation.}

{A. Detection of the group identifier alone: This operation takes a search of $\mathcal{O}(M)$, but is vulnerable to both removal and forgery attempts, as we use a more vulnerable watermark for group identifiers.}

{B. Detection of the Fourier pattern, followed by a validation of the exact initial noise (WIND$_{\text{fast}}$): within the group. This operation takes $\mathcal{O}(N/M)$ search. It is vulnerable to removal attempts (see \mbox{\Cref{tab:performance}}), but more resilient to forgery attempts.}

\begin{figure}[t!]
    \centering
    \includegraphics[width=\textwidth]{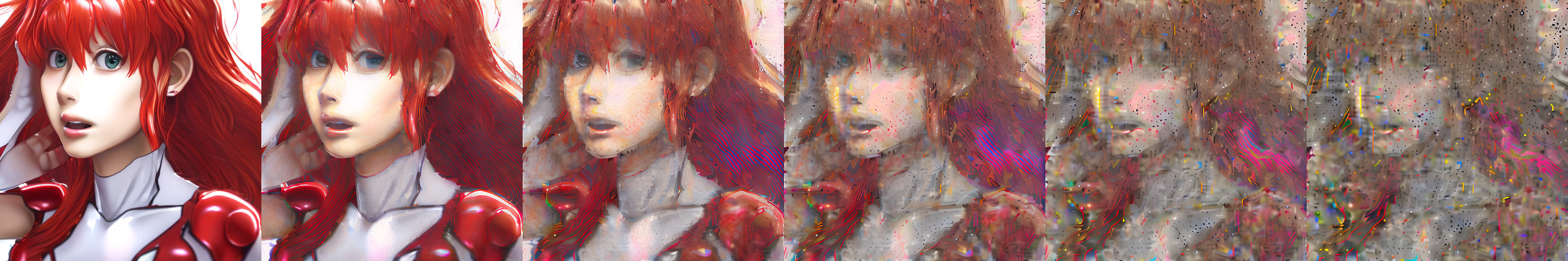}
    \caption{{Image sequence from 0 to 50 regeneration attack iterations.}}
    \label{ab:fig_sequence}
\end{figure}

{C. An exhaustive search of the initial noise, also outside the identified group (WIND$_{\text{full}}$): This operation takes $\mathcal{O}(N)$ search. It is more resilient to both removal and forgery attempts (see \mbox{\Cref{fig:avg_att}}, and \mbox{\Cref{tab:performance}})}.
{This method, while slower is also easier to adapt. A user who wishes to use a fast version of this variant may apply a similar algorithm to the one described above using only a few possible random noises. This would replace the distortion-free property when considering many different images with the ability to rapidly and simply detect the watermarked images. }

{Practically, an NN search can be accelerated using many methods, and can be scaled to tens of millions without significantly affecting the detection time \mbox{\citep{wei2024det, douze2024faiss, malkov2018efficientrobustapproximatenearest, andoni2018approximate}}.}

\paragraph{{{Image Quality Considerations.}}}

{Our method relies on using an initial random noise, drawn from the same distribution of initial noises already used by the model. Therefore, the core of our method (the initial noise stage) is not compromising the visual quality of the generated images at all.}

{The only effect on visual quality comes from the group identifier, where we use existing off-the-shelf watermarking images. In our implementation, we used the RingID~\mbox{\citep{ci2024ringid}} method that adds the Fourier pattern to the initial noise.}

{When a model owner wishes to preserve image quality even better, they may use any other existing watermarking method for the group identifier stage. This will still not compromise the security provided by the random noise seeding stage.}

\paragraph{{{Inversion Attack.}}}
{As discussed in \mbox{\Cref{inv}} in our paper, accurately inverting the model is as difficult as copying the forward process of the model (image generation). While hard, an attacker able to do so is effectively also capable of generating novel images using the same diffusion process. Therefore, At this stage, the model itself is effectively compromised (and not only the watermark signature). We believe that being as hard to forge as the model itself, is a reasonable level of security for almost all use cases. 

Yet, our method does not eliminate the concern that noise could be forged without inverting the model. Approximately inverting the model might also be a threat. While even approximately inverting a model is also hard, it might be easier than stealing the model.  Still, we would like to emphasize that our method is more secure than other diffusion-process-based watermarking techniques, where image distortion themselves may allow easier forging~\mbox{\citep{yang2024steganalysisdigitalwatermarkingdefense}}.}

\section{\texorpdfstring{{Additional Results}}{Additional Results}}
\label{app:additional_results}

\subsection{\texorpdfstring{{Applicability to Other Types of Models}}{Applicability to Other Types of Models}}

{We expect our watermark to be effective directly for any model for which some inversion to the original noise is possible. Namely, as the correlation between random noises in a very high dimension is very much concentrated around 0, even a very slight success in the inversion process is enough to make watermarked images detectable. When considering models with higher generation resolutions, the dimensionality of the noise is even higher, and therefore, we expect the separation would be even better \mbox{\citep{el2009concentration}}.}

{Empirically, to validate the generality of our method, we also report results for the SD 1.4 model \mbox{\citep{rombach2022highresolutionimagesynthesislatent}}. Using $N = 10000$ noises and $M = 2048$ group identifiers, our method achieved a detection accuracy of $\textbf{97\%}$ to identify the correct watermark (initial noise).}

Our method for watermarking the reported SD 2.1 model can also be applied to images obtained from other sources (see \mbox{\Cref{subsec:non-synthetic}}, \mbox{\Cref{app:non-synthetic}}).

\subsection{\texorpdfstring{{Non-Synthetic Images Watermark Detection}}{Non-Synthetic Images Watermark Detection}}
\label{app:non-synthetic}
{Our inpainting method allows us to watermark both images generated by any model and non-synthetic images. To evaluate the robustness of the inpainting watermarking approach for non-synthetic images, we present results similar to those \mbox{\Cref{tab:performance}} for this method, utilizing $N = 100$ initial noises. Results are shown in \mbox{\Cref{tab:ab_tab_inp}}.}

\begin{table}[t!]
    \centering
    \caption{{Inpainting correct watermark detection accuracy.}}
    \begin{tabular}{llccccccccc}
        \toprule
        \textbf{Clean} & \textbf{Rotate} & \textbf{JPEG} & \textbf{C\&S} & \textbf{Blur} & \textbf{Noise} & \textbf{Bright} & \textbf{Avg $\uparrow$} \\ 
        \midrule         1.000&1.000&1.000&0.880&1.000&0.950&0.950&0.969\\
        \bottomrule
    \end{tabular}
    \label{tab:ab_tab_inp}
\end{table}

\subsection{\texorpdfstring{{Further Exploration of the Regeneration Attack Perturbation Strength}}{Further Exploration of the Regeneration Attack Perturbation Strength}}

{In \mbox{\Cref{exp:rob}}, we discussed the robustness of WIND against regeneration attacks. However, using it iteratively might still be a stronger attack against our watermark. We applied the regeneration attack proposed by \mbox{\cite{zhao2023invisibleimagewatermarksprovably}}, up to 50 times. We see that iterative regeneration indeed decreases the similarity between the original initial noise and the reconstructed one. This happens as the image becomes less and less correlated to the original generation \mbox{\Cref{ab:fig_sequence}}.}

\begin{figure}[t!]
    \centering
    \includegraphics[width=0.8\textwidth]{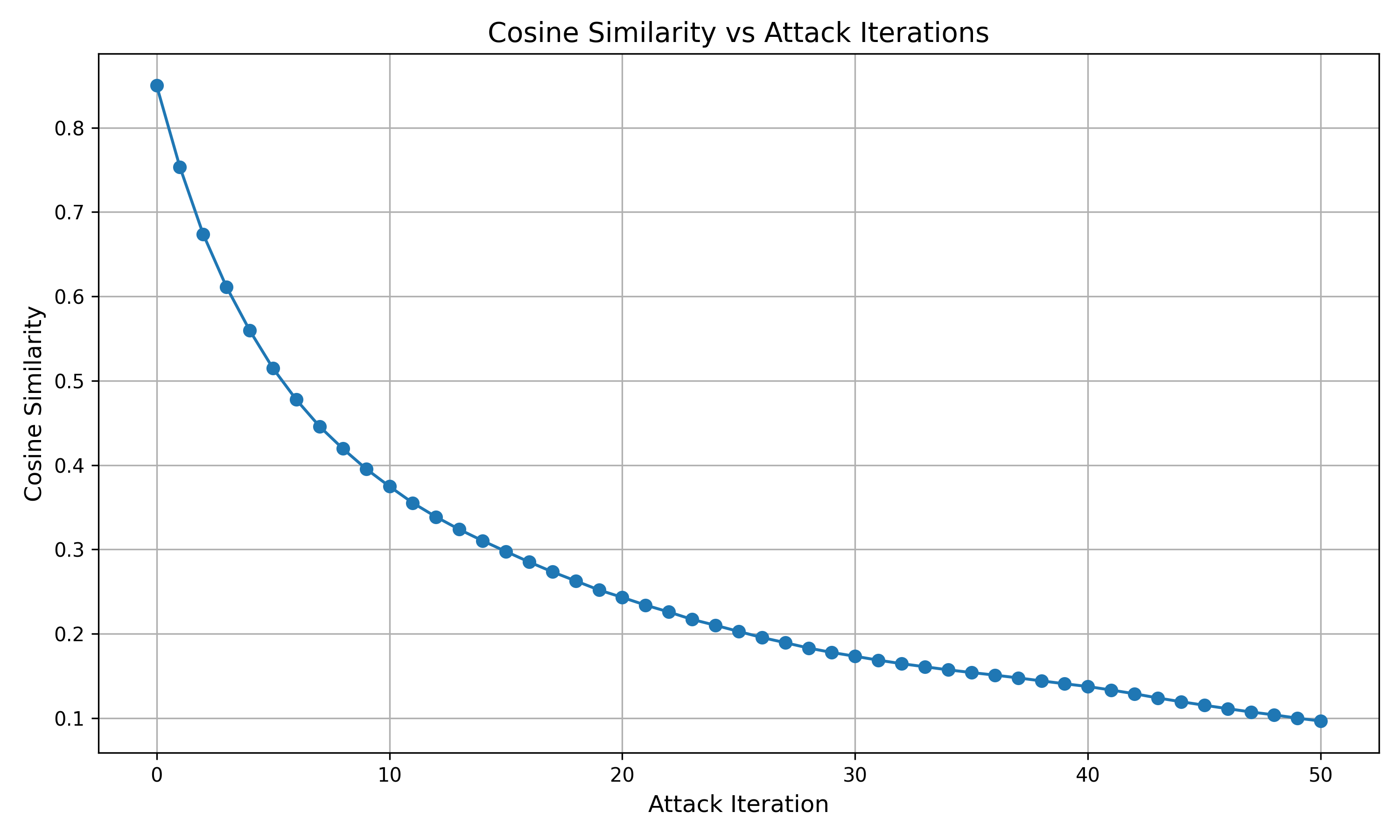}
    \caption{{Cosine Similarity from 0 to 50 regeneration attack iterations.}}
    \label{fig:sequence}
\end{figure}

{Yet, the detection rate of our algorithm remains very high (see \mbox{\Cref{tab:ab_iter}}, \mbox{\Cref{fig:sequence}}). We attribute this to the fact that even a slight remaining correlation between the attacked image and the initial noise remains significant with respect to the correlation expected from non-watermarked images. This happens because of the very low correlation between random initial noises (\mbox{\Cref{fig:true_random}}).}

\begin{table}[h]
    \centering
    \caption{{Correct watermark detection among $10,000$ options after iterative regeneration attack.}}
    \begin{tabular}{lcc}
        \toprule
        \textbf{Iteration} & \textbf{Cosine Similarity} & \textbf{Detection Rate}  \\ 
        \midrule 
        10 & 0.493 & 100\% \\
        20 & 0.342 & 100\% \\
        30 & 0.243 & 100\% \\
        40 & 0.170 & 100\% \\
        50 & 0.121 & 100\% \\
        \bottomrule
    \end{tabular}
    \label{tab:ab_iter}
\end{table}

\subsection{\texorpdfstring{{Quantitative Analysis of the Effect on Image Quality}}{Quantitative Analysis of the Effect of Watermark on Image Quality}}

\begin{table}[t]
    \centering
    \caption{{Effect of WIND on CLIP score.}}
    \begin{tabular}{cc}
        \toprule
        \textbf{CLIP Before Watermark} & \textbf{CLIP After Watermark} \\ 
        \midrule
        0.366 & 0.360 \\
        \bottomrule
    \end{tabular}
    \label{tab:ab_clip}
\end{table}

{We reported the FID of our model on \mbox{\Cref{tab:fid}}. To further assess the effect of WIND watermark on image quality, we report the CLIP score \mbox{\cite{hessel2021clipscore}} before and after watermarking on \mbox{\Cref{tab:ab_clip}}. Our results indicate that adding the watermark has a negligible effect on the CLIP score for generated images.

To further quantify the distortion introduced by each model, we report pixel-base matrices, SSIM and PSNR in the two settings we study:

\mbox{\textbf{Images Generated by the Diffusion Model.}} WIND’s distortion arises from using \textit{group identifiers}, enabling faster detection. To disentangle this effect, we also evaluate WIND$_{\text{w/o}}$, which omits group identifiers. As can be seen in \mbox{\Cref{tab:ab_ssim}}, the image quality generated using our full method is comparable to that of previous techniques. Users who wish to generate distortion-free images, without affecting image quality, can do so by omitting the group identifier (at the cost of a slower detection phase for very large values of $N$). 

\mbox{\textbf{Watermarking Non-Synthetic Images.}}
Additionally, we present results for WIND$_{\text{inpainting}}$, our inpainting-based approach capable of watermarking both non-synthetic images and outputs from other generative models (\mbox{\Cref{tab:ab_ssim_non}}). Although other watermarking methods may better preserve image quality, our image quality remains high. Importantly, to the best of our knowledge, our approach is the only one capable of watermarking non-synthetic images while remaining robust against the regeneration attack~\mbox{\citep{zhao2023invisibleimagewatermarksprovably}}. Therefore, it is preferable when an adversary may try to remove the watermark.

In addition, the inpainting technique can be applied selectively to specific parts of the image if the copyright owner wishes to perfectly preserve fine details in certain areas.

}

\begin{table}[h]
    \centering
    \caption{{SSIM and PSNR values of initial noise-based watermarking approaches. WIND$_{\text{w/o}}$ refers to the method without \textit{group identifiers}}.}
    \begin{tabular}{lcc}
        \toprule
        \textbf{Method} & \textbf{SSIM $\uparrow$} & \textbf{PSNR $\uparrow$} \\ 
        \midrule
        \vspace{0.1cm}
        WIND$_{\text{w/o}}$ & \textbf{1.000} & \textbf{$\infty$} \\
        \vspace{0.1cm}
        WIND$_{\text{full}}$ & 0.494 & 14.647 \\
        \vspace{0.1cm}
        RingID & 0.454 & 13.560 \\
        \vspace{0.1cm}
        Tree-Ring & \underline{0.545} & \underline{15.251} \\
        \bottomrule
    \end{tabular}
    \label{tab:ab_ssim}
\end{table}

\begin{table}
    \centering
    \caption{{SSIM and PSNR values for non-synthetic image watermarking approaches.}}
    \begin{tabular}{lcc}
        \toprule
        \textbf{Method} & \textbf{SSIM $\uparrow$} & \textbf{PSNR $\uparrow$} \\ 
        \midrule
        \vspace{0.1cm}
        WIND$_{\text{inpainting}}$ & 0.768 & 26.806 \\
        \vspace{0.1cm}
        DwtDctSvd & \underline{0.983} & 39.381 \\
        \vspace{0.1cm}
        RivaGAN & 0.978 & \underline{40.550} \\
        \vspace{0.1cm}
        SSL & \textbf{0.984} & \underline{41.795} \\
        StegaStamp & 0.911 & 28.503 \\
        \bottomrule
    \end{tabular}
    \label{tab:ab_ssim_non}
\end{table}

\subsection{\texorpdfstring{{Robustness Comparison to Different Number of Inference Steps}}{Robustness Comparison to Different Number of Inference Steps}}

{We evaluate the impact of inference steps on detection accuracy, as shown in \mbox{\Cref{tab:ab_inf}}. The results indicate that our method is robust to varying the number of inference steps used.}

\begin{table}[t!]
    \centering
    \caption{{Effect of different numbers of inference steps on detection accuracy.}}
    \begin{tabular}{llccccccccc}
        \toprule
        \textbf{Steps} & \textbf{Clean} & \textbf{Rotate} & \textbf{JPEG} & \textbf{C\&S} & \textbf{Blur} & \textbf{Noise} & \textbf{Bright} & \textbf{Avg $\uparrow$} \\ 
        \midrule
        20 &\textbf{1.000}&0.780&\textbf{1.000}&0.880&0.920&\textbf{1.000}&0.960&0.934\\
        50 &\textbf{1.000}&\textbf{0.930}&\textbf{1.000}&\textbf{0.940}&\textbf{1.000}&0.980&0.980&0.976\\
        100 &\textbf{1.000}&\textbf{0.930}&\textbf{1.000}&\textbf{0.940}&\textbf{1.000}&\textbf{1.000}&0.990&\textbf{0.980}\\
        200 &\textbf{1.000}&0.850&\textbf{1.000}&\textbf{0.940}&\textbf{1.000}&\textbf{1.000}&\textbf{1.000}&0.970\\
        \bottomrule
    \end{tabular}
    \label{tab:ab_inf}
\end{table}

\subsection{\texorpdfstring{{True Positive and AUC}}{True Positive and ROC AUC}}

{Expanding on the detection assessment settings discussed in \mbox{\Cref{sec:experiments}}, we reported additional metrics for WIND's ability to detect if an image is watermarked. AUC and True Positive (TPR@1\%FPR) results are available on \mbox{\Cref{tab:ab_tp_auc}}, demonstrating WIND's strong performance, and emphasizing its robustness and reliability. }

\subsection{\texorpdfstring{{Evaluation Against Additional Attacks}}{Evaluation Against Additional Attacks}}

{We evaluate WIND against a diverse set of attacks, including transfer-based, query-based, and white-box attack methods. Specifically, we employ the WeVade white-box attack \mbox{\citep{jiang2023evading}}, the transfer attack described in \mbox{\cite{hu2024transfer}}, a black-box attack utilizing NES queries \mbox{\citep{ilyas2018black}}, and a random search approach discussed in \mbox{\cite{andriushchenko2024jailbreaking}}, adopted to attempt watermark removal. The success rates of these attacks are detailed in \mbox{\Cref{tab:ab_new_atts}}. Notably, none of these methods succeed against WIND, as the correct watermark remains detectable in over 97\% of cases even after applying these attacks.}

\begin{table}[t]
    \centering
    \caption{{Additional watermark detection results of WIND.}}
    \begin{tabular}{cc}
        \toprule
        \textbf{AUC} & \textbf{TP@1\%} \\ 
        \midrule
        0.971 & 1.000 \\
        \bottomrule
    \end{tabular}
    \label{tab:ab_tp_auc}
\end{table}

\begin{table}[t]
    \centering
    \caption{{Success rate of detecting the correct noise among $10,000$ options while withstanding additional attacks.}}
    \begin{tabular}{cccc}
        \toprule
        \textbf{WeVade} & \textbf{Random Search} & \textbf{Transfer Attack} & \textbf{NES Query}  \\ 
        \midrule 
        1\% & 2\% & 3\% & 2\% \\
        \bottomrule
    \end{tabular}
    \label{tab:ab_new_atts}
\end{table}


\subsection{\texorpdfstring{{Attack With a Public Model}}{Attack With a Public Model}}

We report results to a noise reconstruction attack with a public model in \Cref{tab:inversion_att}.

\subsection{\texorpdfstring{{Additional Watermarking Methods}}{Additional Watermarking Methods}}

We provide a comparison to additional watermarking methods in \Cref{tab:ab_performance}.

\begin{table}[h]
    \caption{{Comparison of correct watermark detection accuracy between WIND and previous image watermarking approaches under various image transformation attacks. $\text{WIND}_M$ denotes the use of $M$ groups, with the total number of noises ($N$) specified in the \enquote{Keys} column.}}
    \begin{tabular}{@{}llccccccccc@{}}
        \toprule
        \textbf{Method} & \textbf{Keys} & \textbf{Clean} & \textbf{Rotate} & \textbf{JPEG} & \textbf{C\&S} & \textbf{Blur} & \textbf{Noise} & \textbf{Bright} & \textbf{Avg $\uparrow$} \\ 
        \midrule
        DwtDct & 1 & 0.974 & 0.596 & 0.492 & 0.640 & 0.503 & 0.293 & 0.519 & 0.574 \\ 
        DwtDctSvd & 1 & \textbf{1.000} & 0.431 & 0.753 & 0.511 & 0.979 & 0.706 & 0.517 & 0.702  \\
        RivaGan & 1 & 0.999 & 0.173 & 0.981 & \textbf{0.999} & 0.974 & 0.888 & 0.963 & 0.854 \\
        \midrule
        \multirow{3}{*}{Tree-Ring} 
            &  32    & 0.790 & 0.020 & 0.420 & 0.040 & 0.610 & 0.530 & 0.420 & 0.404 \\ 
            &  128   & 0.450 & 0.010 & 0.120 & 0.020 & 0.280 & 0.230 & 0.170 & 0.183 \\ 
            &  2048  & 0.200 & 0.000 & 0.040 & 0.000 & 0.090 & 0.070 & 0.060 & 0.066 \\ 
        \midrule
        \multirow{3}{*}{RingID} 
            &  32      & \textbf{1.000} & \textbf{1.000} & \textbf{1.000} & 0.530 & 0.990 & \textbf{1.000} & 0.960 & 0.926 \\ 
            & 128     & \textbf{1.000} & 0.980 & \textbf{1.000} & 0.280 & 0.980 & \textbf{1.000} & 0.940 & 0.883 \\ 
            &  2048    & \textbf{1.000} & 0.860 & \textbf{1.000} & 0.080 & 0.970 & 0.950 & 0.870 & 0.819 \\ 
        \midrule
        \multirow{1}{*}{WIND$_{\text{fast}_{128}}$} 
            & 100000 & \textbf{1.000} & 0.780 & \textbf{1.000} & 0.470 & \textbf{1.000} & \textbf{1.000} & 0.960 & 0.887 \\ 
        WIND$_{\text{fast}_{2048}}$ & 100000& \textbf{1.000} &0.870&0.960&0.060&0.960&0.950&0.900&0.814 \\ 
        \midrule
        \multirow{1}{*}{WIND$_{\text{full}_{128}}$} 
            &  100000 & \textbf{1.000} & 0.780 & \textbf{1.000} & 0.850 & \textbf{1.000} & \textbf{1.000} & \textbf{1.000} & \underline{0.947} \\ 
        WIND$_{\text{full}_{2048}}$ & 100000& \textbf{1.000} &0.880&\textbf{1.000}&0.930&\textbf{1.000}&0.990&0.980 &\textbf{0.969} \\ 
        \bottomrule
    \end{tabular}
    \label{tab:ab_performance}
\end{table}

\section{Proof of Resilience to Forgery}\label{app:crypto_thm}

{The WIND method is an approach for generating multiple watermarked images. \mbox{\Cref{thm:crypto_main}} tells us that compromising one or more watermarked images does not give away any information about any other watermarked images. E.g., the adversary cannot ``generate valid reconstructed noise for any other initial noise index $j \neq i$". That said, \mbox{\Cref{thm:crypto_main}} does leave open the possibility that an adversary can take a watermarked image, reconstruct the initial noise only for that image, and use it to attack the method.}

\paragraph{Cryptographic Background}
Consider a cryptographic hash function \texttt{hash}$: \{0,1\}^* \to \{0,1\}^\ell$ with $\ell$ output bits. E.g., $\ell=256$ for SHA-256.
We will describe properties of the hash function in terms of `difficulty'; we say a task is `difficult' if, as far as we know, finding a solution is almost certainly beyond the computational capabilities of any reasonable adversary.
An \textit{unbroken} cryptographic hash function satisfies the following properties:
\textit{Pre-image resistance} requires that given a hashed value $v$, it is difficult to find any message $m$ such that $v = $ \texttt{hash}$(m)$.
\textit{Second pre-image resistance} requires that given an input $m_1$, it is difficult to find a different input $m_2$ such that \texttt{hash}$(m_1) $ = \texttt{hash}$(m_2)$.
\textit{Collision resistance} requires that it is difficult to find two different messages $m_1$ and $m_2$ such that \texttt{hash}$(m_1) $ = \texttt{hash}$(m_2)$.

\cryptothm*

\begin{proof}[Proof of \Cref{thm:crypto_main}]
We will prove each part of the theorem separately:

{1. The adversary cannot recover the secret salt $s$:}
Given the output $\textrm{seed} =$ \texttt{hash}$(i^*, s)$ and partial input $i^*$ the adversary aims to find $s$. This is equivalent to finding a pre-image given partial information about the input. By the pre-image resistance property of cryptographic hash functions, this task is computationally infeasible.
Even if the adversary knows the value of $i$, the space of possible secret salts $s$ is too large to search exhaustively (as $s$ is a sufficiently long random string). Therefore, the adversary cannot recover $s$.

2. The adversary cannot generate valid reconstructed noise for any other initial noise index $j \neq i$.
This security guarantee is ensured by two properties of \texttt{hash}:
a) Second pre-image resistance: Given $(i^*, s)$, it's computationally infeasible to find $(i', s)$ where $i' \neq i^*$ such that \texttt{hash}$(i^*, s) =$ \texttt{hash}$(i', s)$.
b) Collision resistance: It's computationally infeasible to find any two distinct inputs that hash to the same output.
These properties ensure that the adversary cannot find alternative inputs that produce the same hash output, and thus cannot generate valid reconstructed noise for different index numbers $j$.
\end{proof}

 \Cref{thm:crypto_main} leaves open the possibility that an adversary can recover the noise from a watermarked image and use \textit{that} noise to forge a new watermarked image.
We empirically show that a specific implementation of this attack fails without access to the weights of the private diffusion model.

\section{Further Discussion on Distortion}
\label{app:distortion_free}

Using the same initial noise for multiple generations is not distortion-free when examining groups of images. For example, all images with the same prompt $p$ and the same initial noise $z$ will be identical, distorted away from the distribution of groups of images generated with i.i.d noises. Luckily, the huge gap between the similarities distribution of (i) reconstructed vs. used noise, and (ii) reconstructed vs. unrelated noise, allows us to use as many different noise patterns, while still keeping the noise we used more similar to the reconstructed noises compared to unrelated ones. Therefore, limiting the level of distortion in practice.


\begin{table}[h]
    \centering
    \caption{{Detection time (second)}}
    \begin{tabular}{ccc}
        \toprule
        \textbf{WIND} & \textbf{Tree-Ring} & \textbf{RingID}  \\ 
        \midrule 
        22 & 20 & 14 \\
        \bottomrule
    \end{tabular}
    \label{tab:ab_comp_rn}
\end{table}

\section{Empirical Runtime Analysis}
\label{ab:emp}
The runtime of our method is highly sensitive to the available computational resources. To provide a practical estimate, we measured the detection time using a single NVIDIA GeForce RTX 3090. Specifically, we divided 100,000 initial noise samples into 32 groups and reported the detection time. Under these conditions, without special optimizations, the detection phase for 100,000 noise samples takes approximately 22 seconds per detection. {We include a comparison with other methods in \mbox{\Cref{tab:ab_comp_rn}}}.

\section{Implementation Details}
\label{sec:implementation-details}

\paragraph{Diffusion Model.} We use Stable Diffusion 2.1~\citep{rombach2022highresolutionimagesynthesislatent}.

\paragraph{Prompts.} we used the set of prompts from~\cite{gustavosta2024pro}.

\paragraph{Threshold for Detection.} For the first variant \textit{\textbf{WIND$_{\text{fast}}$}} (see \Cref{sec:method}) we use a threshold of min $\ell_2 $ norm $ > 160$ in the first stage and select the best-matching noise in the second stage. The second variant (\textit{\textbf{WIND$_{\text{full}}$}}) does not use a threshold, but rather chooses the noise pattern within the group that has the lowest $\ell_2$ as our candidates for the identified noise. The ability to identify the correct noise among $N$ candidates can be interpreted as achieving a $p$ value of at least $1/N$ in cases where detection was successful.

\paragraph{Retrieval Details During Detection.} We included simple rotation (using intervals of $2$ degrees) and sliding window (window size of 32, stride of 8) searches as part of the retrieval process. These searches do not involve directly optimizing for the specific degrees of rotation or cropping used as attacks. %

\clearpage
\section{Additional Qualitative Results}
\label{appendix:generated_images}

\vspace*{\fill}
\begin{center}
\begin{minipage}[c]{\textwidth}
  \centering
  \includegraphics[width=0.9\textwidth]{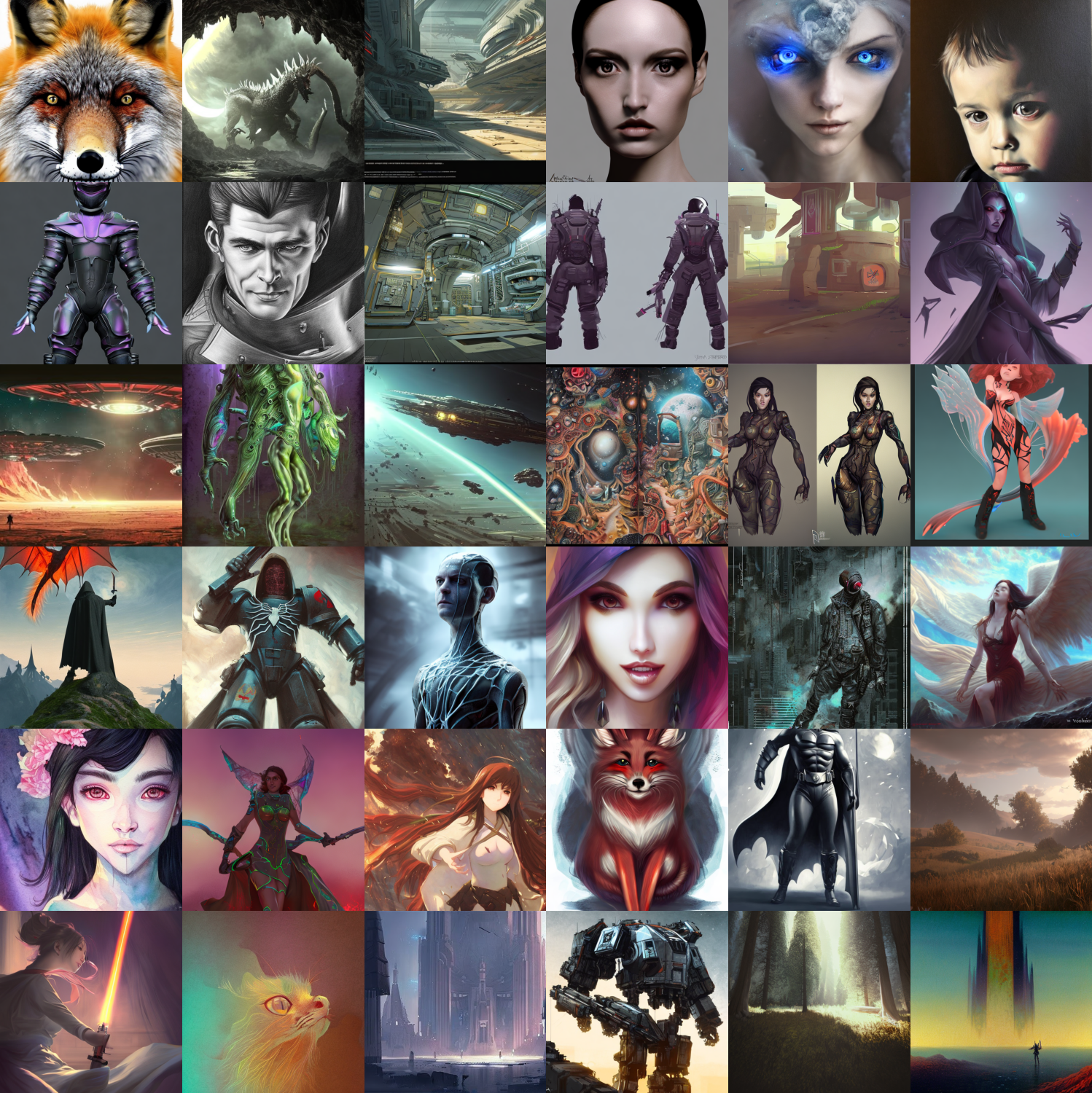}
  \captionof{figure}{More watermarked images generated with WIND.}
  \label{fig:generated_images_grid_1}
\end{minipage}
\end{center}
\vspace*{\fill}
\clearpage

\vspace*{\fill}
\begin{center}
\begin{minipage}[c]{\textwidth}
  \centering
  \includegraphics[width=0.9\textwidth]{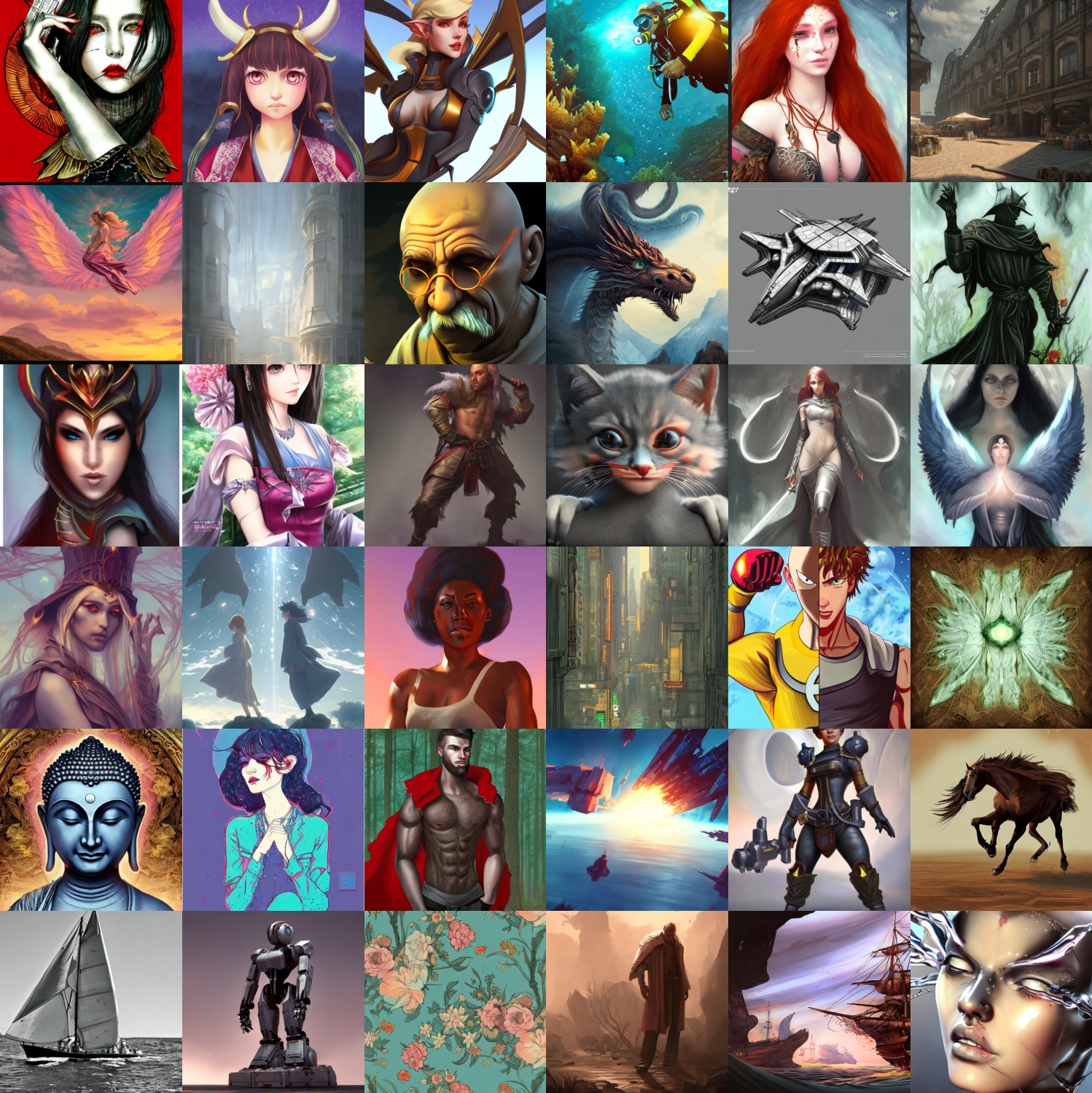}
  \captionof{figure}{More watermarked images generated with WIND.}
  \label{fig:generated_images_grid_2}
\end{minipage}
\end{center}
\vspace*{\fill}
\clearpage

\vspace*{\fill}
\begin{figure}[H] 
    \centering
    \begin{tabular}{cccc}
        \textbf{Before} & \textbf{After} & \textbf{Before} & \textbf{After} \\[6pt]
        
        \includegraphics[width=0.22\textwidth]{images/fig_inp_app/org1.png} &
        \includegraphics[width=0.22\textwidth]{images/fig_inp_app/wm1.png} &
        \includegraphics[width=0.22\textwidth]{images/fig_inp_app/org4.png} &
        \includegraphics[width=0.22\textwidth]{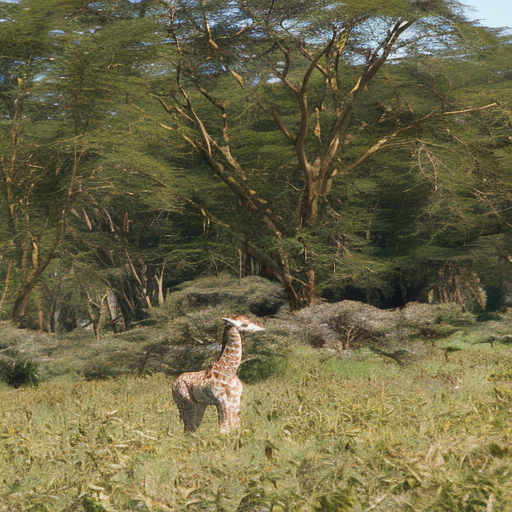} \\[12pt]
        
        \includegraphics[width=0.22\textwidth]{images/fig_inp_app/org3.png} &
        \includegraphics[width=0.22\textwidth]{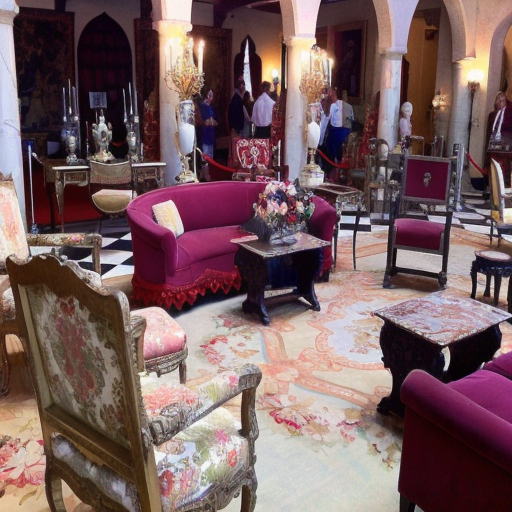} &
        \includegraphics[width=0.22\textwidth]{images/fig_inp_app/org2.png} &
        \includegraphics[width=0.22\textwidth]{images/fig_inp_app/wm2.png} \\[12pt]
        
        \includegraphics[width=0.22\textwidth]{images/fig_inp_app/org5.png} &
        \includegraphics[width=0.22\textwidth]{images/fig_inp_app/wm5.png} &
        \includegraphics[width=0.22\textwidth]{images/fig_inp_app/org6.png} &
        \includegraphics[width=0.22\textwidth]{images/fig_inp_app/wm6.png} \\[12pt]
        
        \includegraphics[width=0.22\textwidth]{images/fig_inp_app/org7.png} &
        \includegraphics[width=0.22\textwidth]{images/fig_inp_app/wm7.png} &
        \includegraphics[width=0.22\textwidth]{images/fig_inp_app/org8.png} &
        \includegraphics[width=0.22\textwidth]{images/fig_inp_app/wm8.png} \\
    \end{tabular}
    \caption{More comparisons of COCO images before and after watermarking with WIND.}
    \label{app:watermark_comparison}
\end{figure}
\vspace*{\fill}

\clearpage

\vspace*{\fill}
\begin{center}
\begin{minipage}[c]{\textwidth}
  \centering
  \includegraphics[width=0.9\textwidth]{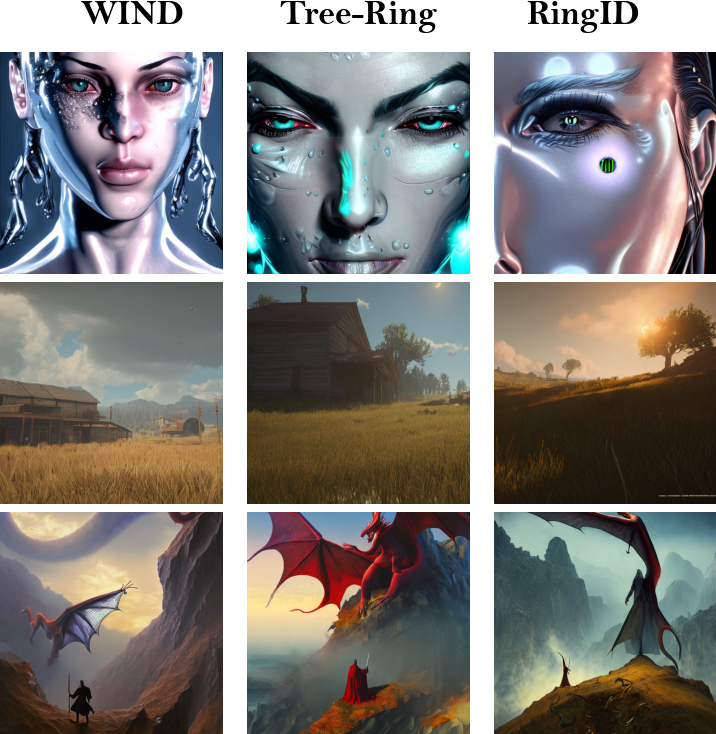}
  \captionof{figure}{{More qualitative results of watermarked images generated using WIND, Tree-Ring, and RingID.}}
  \label{fig:generated_images_grid_2}
\end{minipage}
\end{center}
\vspace*{\fill}
\clearpage
\clearpage

\vspace*{\fill}
\begin{center}
\begin{minipage}[c]{\textwidth}
  \centering
  \includegraphics[width=0.9\textwidth]{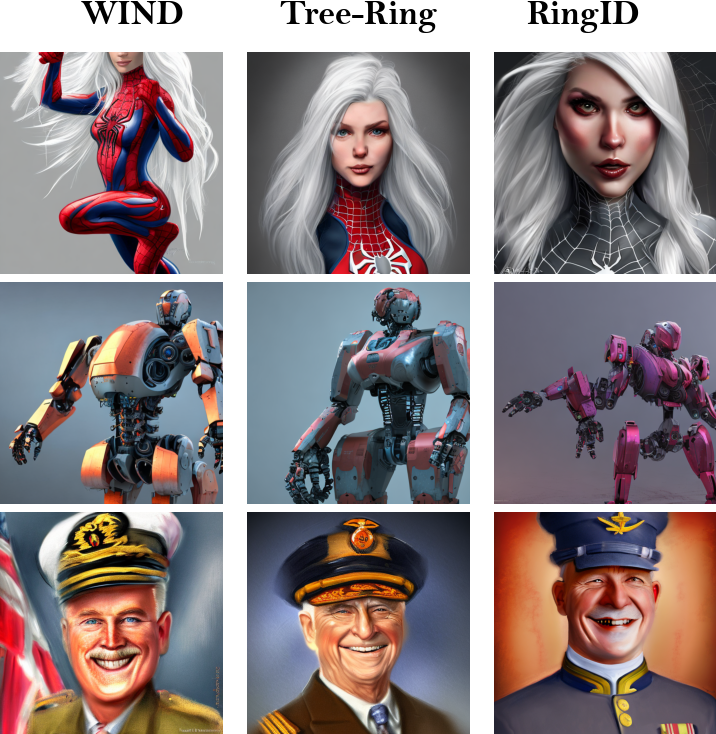}
  \captionof{figure}{{More qualitative results of watermarked images generated using WIND, Tree-Ring, and RingID.}}
  \label{fig:generated_images_grid_2}
\end{minipage}
\end{center}
\vspace*{\fill}
\clearpage

\clearpage

\vspace*{\fill}
\begin{center}
\begin{minipage}[c]{\textwidth}
  \centering
  \includegraphics[width=0.9\textwidth]{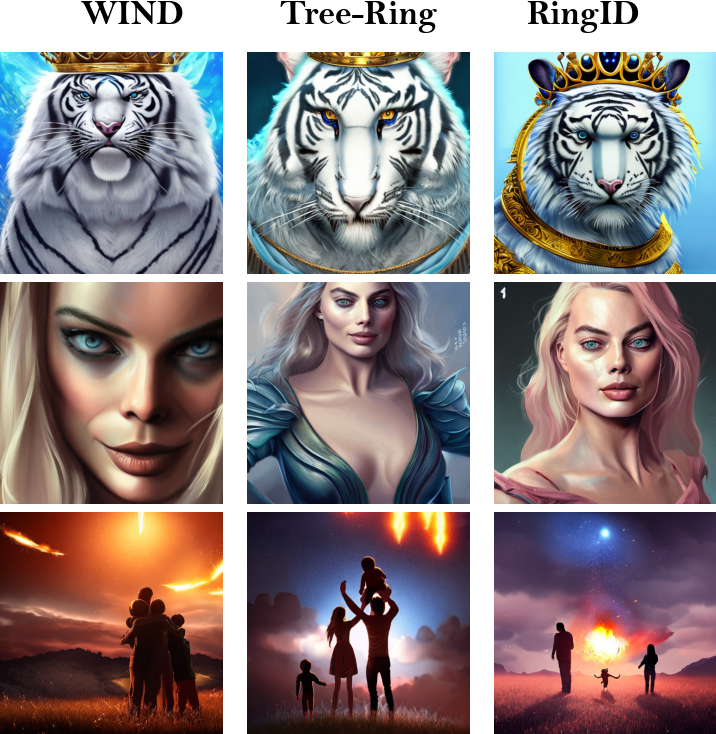}
  \captionof{figure}{{More qualitative results of watermarked images generated using WIND, Tree-Ring, and RingID.}}
  \label{fig:generated_images_grid_2}
\end{minipage}
\end{center}
\vspace*{\fill}
\clearpage

\clearpage

\vspace*{\fill}
\begin{center}
\begin{minipage}[c]{\textwidth}
  \centering
  \includegraphics[width=0.9\textwidth]{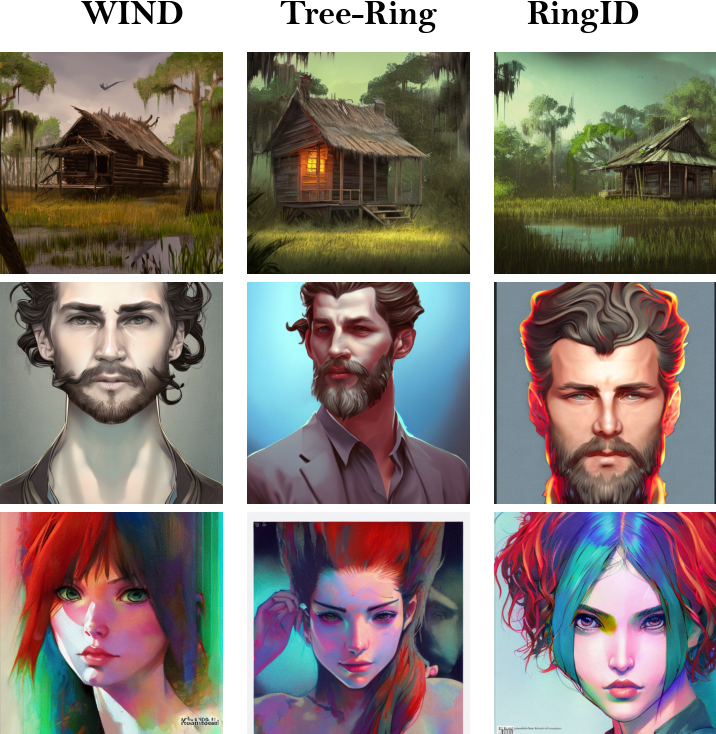}
  \captionof{figure}{{More qualitative results of watermarked images generated using WIND, Tree-Ring, and RingID.}}
  \label{fig:generated_images_grid_2}
\end{minipage}
\end{center}
\vspace*{\fill}
\clearpage

\clearpage

\vspace*{\fill}
\begin{center}
\begin{minipage}[c]{\textwidth}
  \centering
  \includegraphics[width=0.9\textwidth]{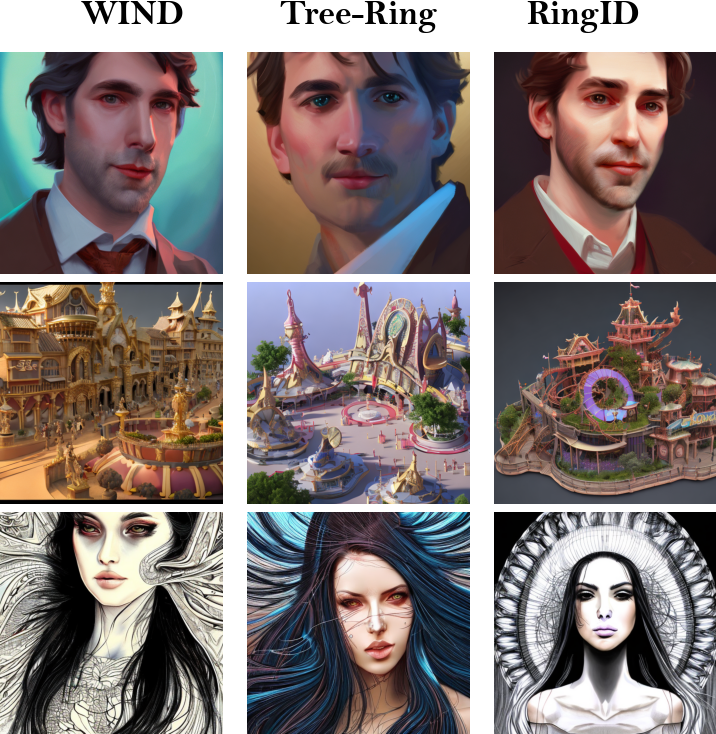}
  \captionof{figure}{{More qualitative results of watermarked images generated using WIND, Tree-Ring, and RingID.}}
  \label{fig:generated_images_grid_2}
\end{minipage}
\end{center}
\vspace*{\fill}

\end{document}